\documentclass[10pt,twocolumn,letterpaper]{article}

\usepackage[pagenumbers]{iccv} 

\definecolor{iccvblue}{rgb}{0.21,0.49,0.74}
\usepackage[pagebackref,breaklinks,colorlinks,allcolors=iccvblue]{hyperref}
\usepackage{multicol}

\newtheorem{theorem}{Theorem}[section]
\newtheorem{proposition}[theorem]{Proposition}

\newtheorem{definition}[theorem]{Definition}

\def\paperID{*****} 
\def\confName{ICCV}
\def\confYear{2025}

\title{BACK TO THE FUTURE CYCLOPEAN STEREO:\\A Human Perception Approach Combining Deep and Geometric Constraints}

\author{
    Sherlon Almeida da Silva¹²
\and
    Davi Geiger²
\and
    Luiz Velho³
\and
    Moacir Antonelli Ponti¹
\and
    {\normalsize ¹Instituto de Ciências Matemáticas e de Computação, Universidade de São Paulo (ICMC-USP)} \\
    {\normalsize ²Courant Institute of Mathematical Sciences, New York University (NYU)} \\
    {\normalsize ³Instituto de Matemática Pura e Aplicada (IMPA)}
\and
     {\tt\small \{sherlon.a, dg1\}@nyu.edu},
     {\tt\small lvelho@impa.br},
     {\tt\small moacir@icmc.usp.br}
}

\begin{document}
\maketitle
\begin{abstract}
We innovate in stereo vision by explicitly providing analytical 3D surface models as viewed by a cyclopean eye model that incorporate depth discontinuities and occlusions. 
This geometrical foundation combined with learned stereo features allows our system to benefit from the strengths of both approaches. We also invoke a prior monocular model of surfaces to fill in occlusion regions or texture-less regions where data matching is not sufficient.   Our results already are on par with the state-of-the-art purely data-driven methods and are of much better visual quality, emphasizing the importance of the 3D geometrical model to capture critical visual information. Such qualitative improvements may find applicability in virtual reality, for a better human experience, as well as in robotics, for reducing critical errors.  
Our approach aims to demonstrate that understanding and modeling geometrical properties of 3D surfaces is beneficial to computer vision research. 
\end{abstract}    
\begin{figure*}[!ht]
    \centering
    \includegraphics[width=0.8\textwidth]{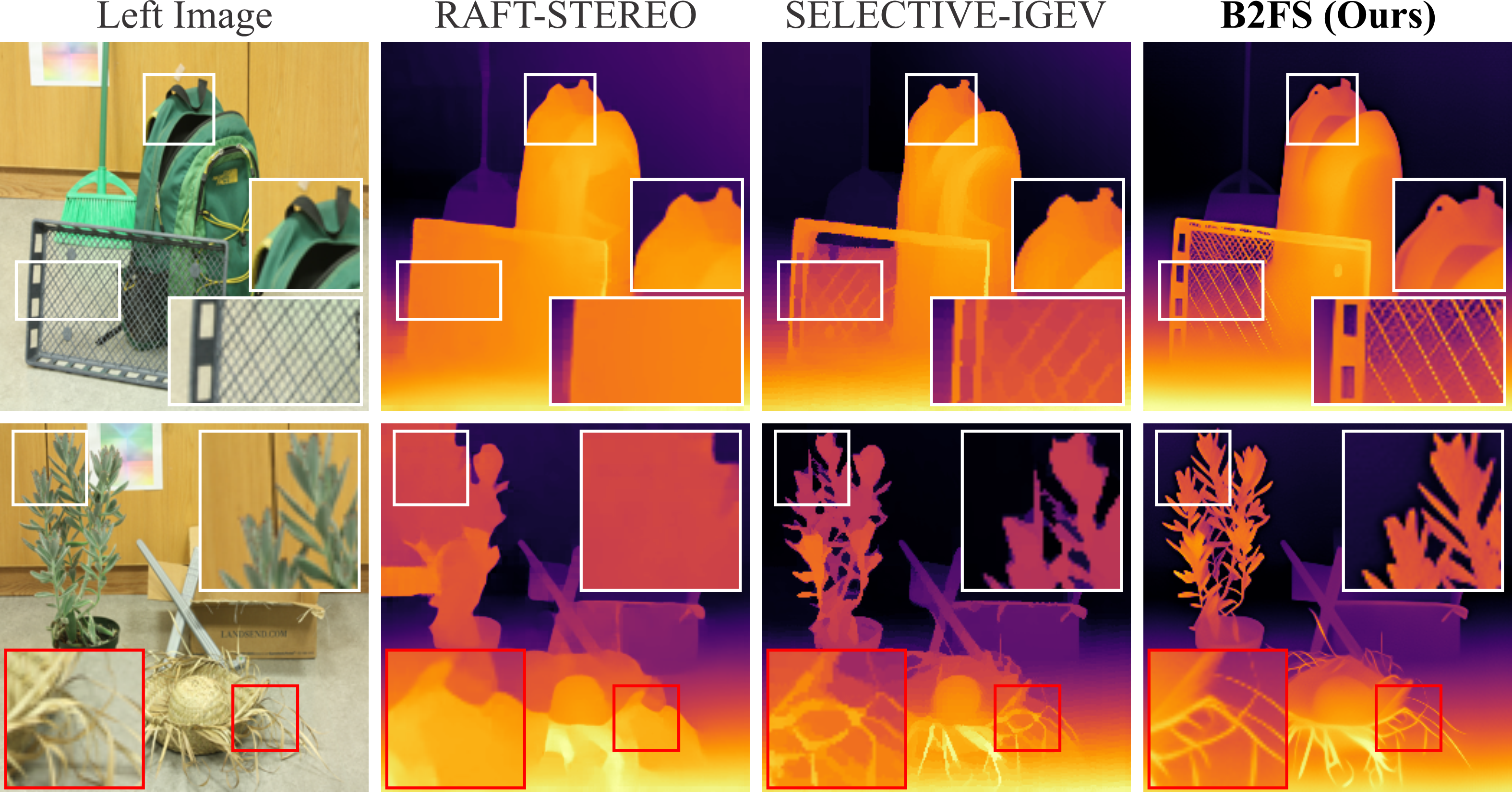}
    \caption{A Comparison of 256x256 results: we present the left image followed by RAFT-Stereo~\cite{lipson2021raft}, Selective-IGEV~\cite{wang2024selective}, and B2FS (Ours). In each image a rectangle is selected and zoomed in (overlayed over the image) to show the visual differences in such areas.}
    \label{fig:techniques_comparison}
\end{figure*}

\section{Introduction}
\label{sec:introduction}

This paper is about stereo vision, and, as in many vision problems, the state-of-the-art for benchmark datasets is obtained via deep learning (DL). For such methods, the input is a stereo pair of images, and training is done with a loss function being the error between the output and the ground truth (GT) disparity. Then, a DL-based stereo model is ready to output disparity data for a new stereo pair of images.
In this context, the more variety and volume of training data, the better the results.  What is concerning is that, despite good performance, it remains a ``black box'', and we (humans) do not understand exactly how it comes to the results. 

What does it mean to ``understand stereo vision'' and why does it matter? Understanding means discovering internal representations of the problem at hand, and often it is needed to generalize to new situations or to allow us to apply such knowledge broadly. Suppose that a camera records video data of objects falling to the ground. It is plausible that an algorithm from machine learning (ML) could make predictions of such trajectories with an accuracy equal to or greater than any physics model, assuming enough variety and quantity of data are provided. However, such an ML solution would not have revealed the underlying laws of motion that govern them, such as the formula $F=m a$ and $a=9.8 m/s^2$, an abstraction of reality that alone does not even give the most accurate models when the wind plays a role in the trajectory. Still, it has been a key insight for engineering and our civilization. Our goal is to better understand stereo vision with the mandate that to demonstrate better understanding, the new solution must perform on par or better with respect to the state-of-the-art stereo vision. In this way, the abstractions of the 3D reality we make here can help not only to improve efficiency and performance of stereo algorithms, but perhaps such 3D models of reality can be used for other activities in computer vision.

Furthermore,  ML extracts features that enable abstractions of the visual world to be employed and tested. This paper does not claim to have fully understood 3D reality, but instead that some significant understanding of it is made to help produce solutions on par with the state-of-the-art (pure data-driven) stereo algorithms and with clear superior depth maps visual appearance as depicted in Figure~\ref{fig:techniques_comparison}.

Let us point out that some of the ideas developed here are rooted in previous stereo studies dating back to the early work of Helmholtz~\cite{Helmholtz1910} and Julesz~\cite{Julesz71} that created and advocated for the cyclopean eye view of stereo vision, and also several computer vision studies before the dominance of pure data-driven DL methods. 


\subsection{Previous Work}

``Back then'' in 1499, in his Treatise on Painting, Leonardo da Vinci explained that different parts of the background are occluded when viewed from the left eye compared to the right eye~\cite{davinci2014treatise}.

Hermann von Helmoltz's \cite{Helmholtz1910} in the late 1800's and early 1900's was a pioneer for pointing out the use of binocular vision to achieve depth perception. Helmholtz conceptualized the ``cyclopean eye'' as a hypothetical single eye located in the middle of the head, representing the brain's integration of the two eyes' input. Bela Julesz pushed the study of stereo vision toward human perception and modeling \cite{Julesz71} while creating the random dot stereogram (showing that stereo works even without recognition). 

Marr and Poggio \cite{MarrPoggio79} attempted to mathematically model human stereo vision, matching zero crossings in filtered images that correspond to edges and features in the visual field, and delivering the $2\frac{1}{2}$ sketch (including the surface reconstruction process). This work was followed by several computational models in the 1980's, 1990's, and 2000's, where in particular global modeling of occlusions and discontinuities with dynamic programming (DP) optimization \cite{Geigeretal92, Geigeretal95,Belhumeur96} and with the minimum cut optimization \cite{IshikawaGeiger98,Boykov2001}, and all their citations, were the dominant approach to stereo.   
In all these models, relatively little progress was achieved in improving feature extraction for matching, though there were efforts, e.g. with an overcomplete set of linear filters~\cite{MalikJones92} and with left and right windows per image pixel to compete for matching~\cite{Geigeretal92}. Extraction of local $L-Y-T-$junctions to infer depth ordering in monocular images also did not progress much. 

With the advent of DL and the availability of datasets, a much better level of accuracy has been achieved in stereo by RAFT-Stereo~\cite{lipson2021raft}, CREStereo~\cite{li2022practical}, DLNR~\cite{zhao2023high}, and Selective-IGEV~\cite{wang2024selective}. These techniques learn to produce disparity maps with performance higher than that of all previous stereo algorithms. In close examination, these methods first extract features from the left and right images and later perform disparity estimation at each pixel. 

More recently, advances in qualitative depth recovery from the monocular view, through DL techniques and large datasets, have produced excellent results, e.g. MiDaS~\cite{ranftl2020towards}, DPT~\cite{ranftl2021vision}, Metric3D~\cite{yin2023metric3d, hu2024metric3d}, ZoeDepth~\cite{bhat2023zoedepth}, PatchFusion~\cite{li2024patchfusion}, UniDepth~\cite{piccinelli2024unidepth}, Marigold~\cite{ke2024repurposing}, Depth Anything~\cite{yang2024depthv1, yang2024depthv2}, and DepthPro~\cite{bochkovskii2024depth}. Upon reflection, these methods likely have excellent implicit surface models.

\subsection{Our Contribution}
Broadly presented, here are our main contributions.

\noindent-- \textbf{Hybrid Geometric–Learning Framework:} a novel stereo vision approach that combines geometric reasoning with the adaptability of DL (see Figure~\ref{fig:b2fs_workflow}).

\noindent-- \textbf{Cyclopean Eye Model:} a novel framework that extracts depth discontinuities and occlusions, imposing a human-like perception of 3D scenes that for opaque surfaces there is one and only one disparity solution per cyclopean spatial coordinate $(e,x)$ (see Figures~\ref{fig:space_transformation}, \ref{fig:monocular_cyclopean_depth}, and~\ref{fig:cyclopeaneye}). 

\noindent-- \textbf{Occlusion and Texture-Less Region Depth Recovery:} incorporate a prior monocular surface model to fill in the depth values where occluded and texture-deficient regions occur (see Figures~\ref{fig:cyclopeaneye} and \ref{fig:lr_correlation}).

\noindent-- \textbf{Foundational Contribution to Vision Understanding:} deepens our understanding of 3D scene geometry, paving the way for more interpretable, generalizable, and reliable vision systems.

\section{Back to GCs: Occlusions and Discontinuities}
\label{sec:model}
Our approach goes "back" to the cyclopean coordinate system (XD) where the geometric constraints (GCs) of the  3D world can be best described. 

\subsection{Cyclopean Coordinate System (XD)}
\label{sec:cyclopean}
\begin{figure}
    \centering
    \includegraphics[width=\columnwidth]{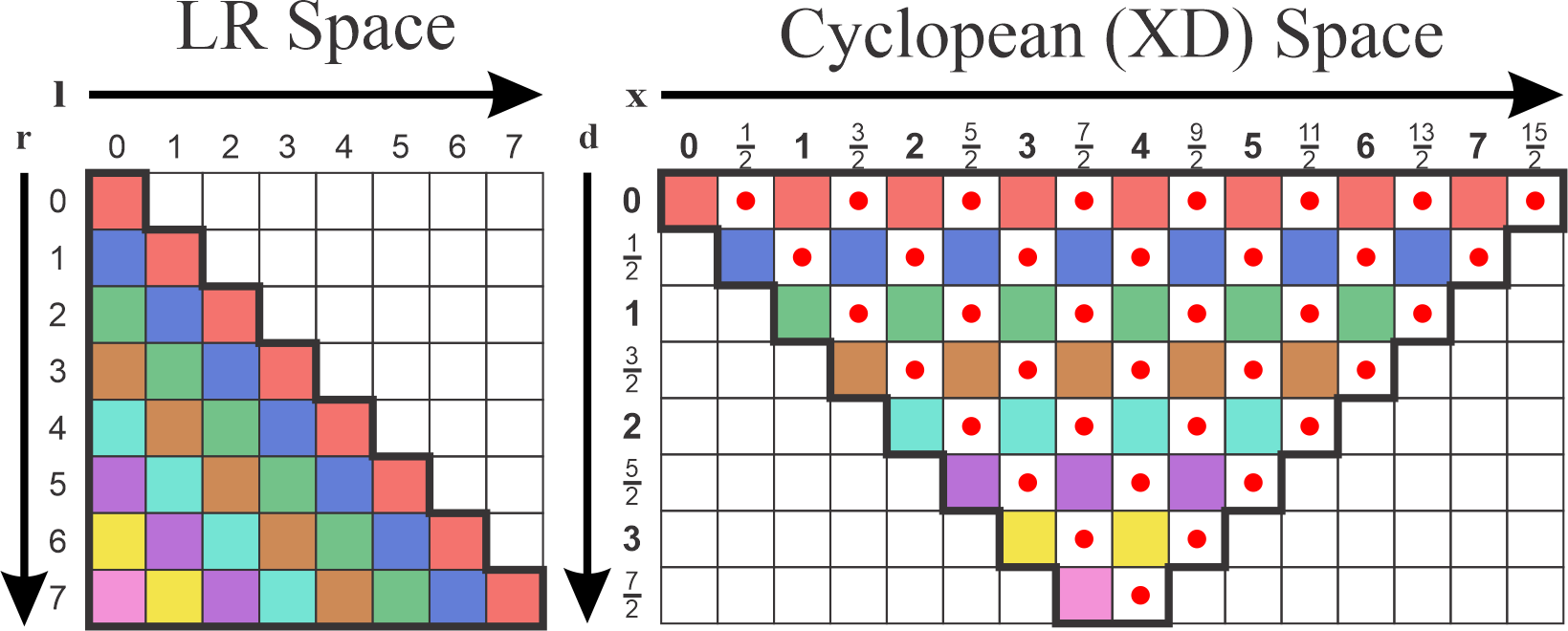}
    \caption{Space Transformation from L$\times$R CS (left) to the XD (right). The colors represent a disparity value. Empty positions in XD space are disallowed, while the 'red dot'  data are obtained via a bilinear interpolation from LR space data. The XD space has twice the resolution of the LR space, for each epipolar line.}
    \label{fig:space_transformation}
\end{figure}

Consider a left image $I^L$ with height and width $M \times  N$ and a pixel coordinate system (CS) $(e,l) \in M \times  N$ and consider the $M 
\times  N$ right image $I^R$ and its CS $(e,r) \in M \times  N$ where $e \in (0,1,\hdots M-1)$ index their respective epipolar lines (see Figure~\ref{fig:space_transformation} and~\ref{fig:cyclopeaneye}). For many datasets, such as Middlebury~\cite{scharstein2014high}, the images have been rectified by the epipolar lines, which are then simply the horizontal lines of the images. The space $(e,x,d) \in M \times 2N \times D$ allows us to describe the matching of $(e,l) \leftrightarrow (e,r)$  as an assignment of a disparity $d$ to $(e,x)$ (see Figure~\ref{fig:space_transformation}). 

Such matching and associated assignment is described by the invertible coordinate transformation:
\begin{align}
    \begin{pmatrix}
        x \\ d
    \end{pmatrix} = \frac{1}{2} \begin{pmatrix}
        1 & 1\\ 1 & -1 
    \end{pmatrix} \begin{pmatrix}
        r \\ l
    \end{pmatrix}\, .
    \label{eq:cyclopean-transformation}
\end{align}
One consequence of this transformation is that the discrete cyclopean width coordinate  $x\in (0, \frac{1}{2}, 1,  \frac{3}{2}, \hdots,  N-1, N-\frac{1}{2})$ has subpixel resolution, twice as much as the image pixel width resolution of $l, r \in (0, 1, \hdots, N-1)$ (see Figure~\ref{fig:space_transformation}). 

The XD provides a depth value ${\cal D}(e,x)$, for each match $(e,l) \leftrightarrow (e,r)$, as follows
\begin{align}
    {\cal D}(e,x) = f \frac{B}{d(e,x)}\, ,
    \label{eq:disparity-depth}
\end{align}
where $f$ is the focal length, $B$ is the baseline (distance between the left and right camera centers), see Figure~\ref{fig:monocular_cyclopean_depth}. The depth provided by man-made datasets is typically obtained with a light projection or laser that simulates the view of the XD, and not the depth from the L or R.

\begin{definition}[Opaque Surfaces and Stereo] 
\label{def:opaque-surfaces}
3D opaque surfaces do not let light pass through them. Consider a 3D point $P=(X,Y,Z)$ that belongs to an opaque surface. If it is visible by L we can describe it as $(e,l, {\cal D}^L(e,l))$ and if it can be seen by R it can be described as $(e,r, {\cal D}^R(e,r))$. 
\end{definition}
Note that (i) if $P$ is visible by both eyes,  still in general ${\cal D}^L(e,l)\ne {\cal D}^R(e,r)$, but the disparities satisfy $d^L(e,l)=d^R(e,r=2d^L(e,l)+l)$; (ii) Also, in general,  given a match $(e,l) \leftrightarrow (e,r)$ and the assignment $d(e,x)$ derived from \eqref{eq:cyclopean-transformation} leading to ${\cal D}(e,x)$, obtained from \eqref{eq:disparity-depth},  we have ${\cal D}(e,x)\ne {\cal D}^L(e,l),  {\cal D}^R(e,r)$.

\begin{definition}[Transparent Surfaces and Stereo] 
\label{def:transparent-surfaces}
Transparent surfaces allow some light to pass through. 
Two distinct points $P_{1,2}=(X_{1,2},Y_{1,2},Z_{1,2})$ are in a transparent pair state if both can be seen by the XD and share the same $(e,x)$ coordinates. Transparent stereo surfaces are a set of contiguous transparent-pair states. 
\end{definition}

As we describe next, the XD allows for a simple description of the geometrical constraints of a stereo vision system. 

\subsection{Depth from L, R, and XD}

\begin{figure}
    \centering
    \includegraphics[width=\columnwidth]{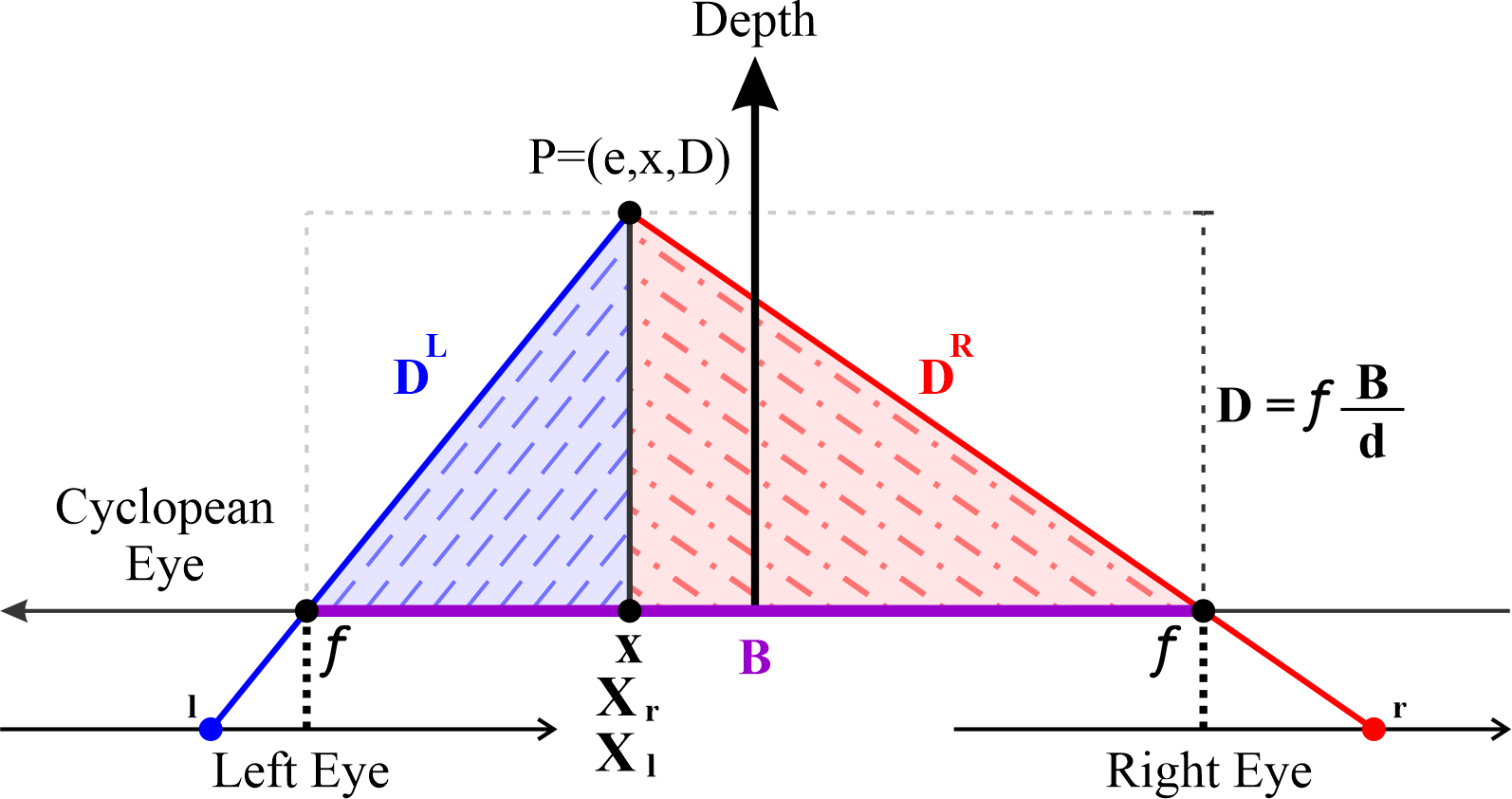}
    \caption{ $D^{L/R}(e,l/r)$ is the depth from L/R CS, respectively.  A point $P$ in 3D is described by the  XD as $P=P^{C}(e, x, Z={\cal D}(e,x ))$. The same point can be described by the L/R CS as $P=P^{L,R}(e,X_{l,r}, {\cal D}(e,x ))$, where $X_{l,r} \ne l,r$, since $l,r$ are the projective projection of $P$ into the L/R CS, while $X_{l,r}$ is the simpler orthogonal projection of $P$ into the L/R CS. Note that $B=X_l-X_r$. The distance to  $P$ measured by the L, R,  and cyclopean eye are ${\cal D}^L(e,l), {\cal D}^R(e,r), {\cal D}(e,x)$, respectively, and they are all different values. Note that the relation $D= f \frac{B}{d}$ assumes $d=r-l$, but our definition of disparity requires a factor 2.  
    }
    \label{fig:monocular_cyclopean_depth}
\end{figure}

Based in Figure~\ref{fig:monocular_cyclopean_depth} we infer the following triangle relations
\begin{align}
\left ( {\cal D}^{L}(e,l) \right)^2 =   {\cal D}^2(e,x)  + \left(\frac{B}{2}-x\right )^2 \quad \textrm{and} \nonumber \\
    \left ( {\cal D}^{R}(e,r) \right)^2 =  {\cal D}^2(e,x)  + \left(\frac{B}{2}+x\right )^2 
    \hspace{0.3in}
    \label{eq:depth-LR}
\end{align}
These relationships introduce a bias in the depth estimation from the L and R cameras relative to the XD depth values. To the best of our knowledge, these have not been derived previously. The impact becomes evident when comparing depth estimates with man-made GT scenarios. A consequence of mismatches in L/R depth estimation yield, for example, motion sickness in virtual reality~\cite{xia2024mismatch}.

\subsection{Occlusions, Discontinuities and GCs}
\label{sec:OcclusionsDiscontinuities}

\begin{definition}[Occlusions and Discontinuities]
R- (L-) occlusions are regions that are seen by the R (L) eye but not by  L (R).  R- (L-) discontinuities are places where jumps of disparity (or of depth) occur in the R (L).  L- and R-occlusions, together as a group, are termed half-occlusions~\cite{Belhumeur96, Wang_2019_CVPR}. 
\end{definition}

As first observed by Da Vinci and recently pointed out in  \cite{Geigeretal92, Geigeretal95, Belhumeur96, IshikawaGeiger98}, 3D opaque surfaces follow geometric constraints that link discontinuities to occlusions. 
\cite{Geigeretal92, Geigeretal95} proposed a monotonicity constraint for the L and R disparity maps for the visible areas for both eyes. 
We next describe our new proposed GCs. 
\begin{proposition}
\label{lemma:Geometrical-Constraint}
GCs for opaque surfaces\\
GC1. The size of the jump, along an epipolar line, of a R- (L-) discontinuity is equal to the size of the L- (R-) occlusion.\\
GC2. Each cyclopean coordinate $(e,x)$ has one and only one disparity, i.e.,  $d$ is a function $d:(e,x) \rightarrow \mathbb{R}$.
\end{proposition}

\begin{figure}
    \centering
    \includegraphics[width=\columnwidth]{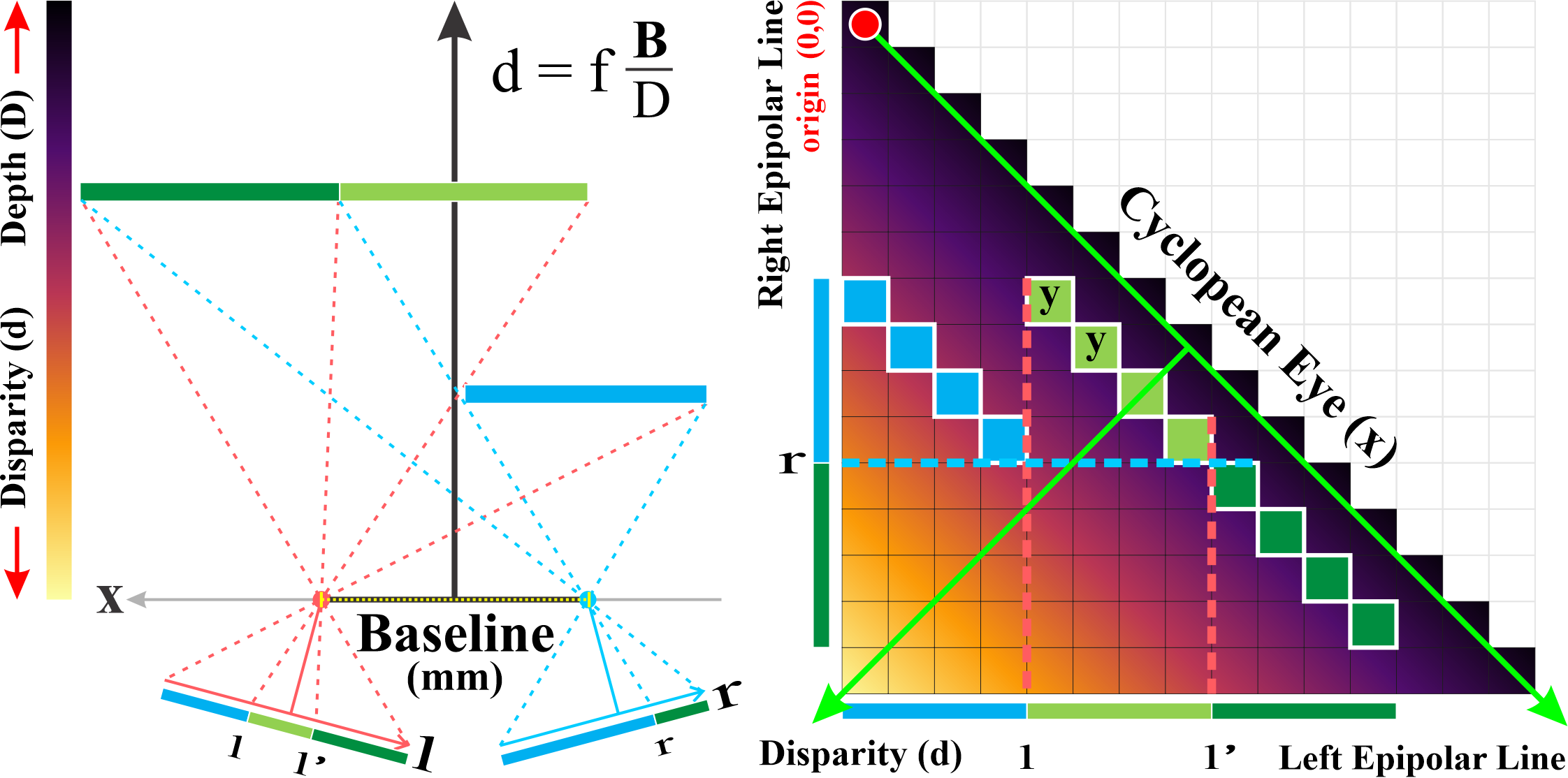}
    \caption{An epipolar slice of a surface with left occlusion region and its description by the XD. a. A top view of the epipolar slice of the surface and the two eyes projections. The baseline ${\bf B}$  connects L to R focal centers. The depth axis describe the inverse of disparity (Equation \eqref{eq:disparity-depth} depicted).  b. A discrete XD, a rotation of the L-R CS described by Equation~\eqref{eq:cyclopean-transformation}. Note that the R CS  is pointing down. The two red vertical dashed lines delimit the L occlusion area which are associated with a R discontinuity along the horizontal blue dashed line with a jump of the same size as the left occlusion, as described by $GC1$ in Proposition~\ref{lemma:Geometrical-Constraint}). Note that the only two (2) light green squares (the ones without an "y" in them) are seen by  the XD associated with the L occlusion, satisfying one disparity per coordinate $x$ (as postulated by $GC2$), which is  {\bf half}  of the size of the L occlusion area. }
    \label{fig:cyclopeaneye}
\end{figure}

Figure~\ref{fig:cyclopeaneye} illustrates these two constraints. To our knowledge, $GC2$ has not been proposed before. Note that even recently Wang~\cite{Wang_2019_CVPR} argued that a unique disparity constraint should not be used for the left eye nor for the right eye, which we agree,  but nothing was mentioned for the cyclopean eye. Clearly, $GC2$ will not be satisfied for transparent surfaces. Although $GC1$ is present in the work \cite{Geigeretal92}, and later, e.g. \cite{Belhumeur96, IshikawaGeiger98}, it has not been precisely modeled in the XD as we do next.  Moreover, the trust of our approach is the integration of GCs with deep learning to create B2FS.
   
Following Proposition~\ref{lemma:Geometrical-Constraint} we model surfaces $\mathcal{S}=\{(e,\mathcal{P}_e);e=0,\hdots, M-1\}$ as a set of paths $\mathcal{P}_e$ in the cyclopean space and indexed by each epipolar line $e$. A path $\mathcal{P}_e=\{[O(e,x), d(e,x)]; \, x=0, 1/2, 1, 3/2, \hdots N-1, N-1/2\}$ is described by a sequence of states $[O(e,x), d(e,x)]$ for each $(e,x)$,  where a binary occlusion variable $O(e,x)=0,1$  and the disparity value $d(e,x)$ are unique (that is, satisfying $GC2$). Here we do not distinguish whether a surface is described in terms of depth or disparity values. In order to impose $GC1$  the path $\mathcal{P}_e$ obeys the following constraint 
{\small
\begin{align}
\textrm{If}\,\,  0=O(e,x)=O(e,x') \quad \textrm{and} \hspace{1.0in}
 \nonumber \\
 1=O(e,x+ 1/2)=O(e,x+1)= \hdots = O(e, x'- 1/2)
 \nonumber \\
\textrm{Then,}\quad |d(e,x')-d(e,x)|=x' - x \hspace{0.8in}
\label{eq:GC1}
\end{align}
}
and if $d(e,x')-d(e,x)>0 $ it is a left occlusion and discontinuity, else if $d(e,x')-d(e,x)< 0 $ it is a right occlusion and discontinuity. 

$GC1$, given by \eqref{eq:GC1}, can be imposed in a local form as 
{\small
\begin{align}
    O(e,x)O\left(e,x-\frac{1}{2}\right)=1  \rightarrow   d(e,x)-d\left(e,x-\frac{1}{2}\right)=\pm \frac{1}{2} \, .
    \label{eq:GC1-DP}
\end{align}
}
Note that the disparity values assigned at such occlusion locations are not part of the final solution, they are just placeholders to ensure that $GC1$ is satisfied. This specific characterization in XD, to the best of our knowledge, has not been proposed before, and it is a critical description to impose $GC1$ locally. Later, the disparity values at the occlusions will have to be estimated. 

In order to find $\mathcal{S}$ given a pair of left and right images, one must define a criterion of what makes $\mathcal{S}$ the correct solution, as well as a method to obtain it (including the constraint \eqref{eq:GC1-DP}). The method we use is DP as we discuss next. 

\subsection{Optimization Criteria and DP}

Along an epipolar line a path must have good matches between left and right image features in regions of no occlusions. Define a measure of Feature Matching Similarity (FMS) as a scalar product between features in L and R, i.e.,  
{\small
\begin{align}
   FMS(e,x,d(e,x))=F^L(e,x-d(e,x))\cdot F^R(e,x+d(e,x))
   \label{eq:fm}
\end{align}
}
and so $FMS_{(e,x)}(d(e,x))$ must be large when there is no occlusion. 
In addition, the regions of the occlusions tend to be contiguous (not isolated points). Figure~\ref{fig:lr_correlation} illustrates the data quality of such matching (a distance $FM_{(e,x)}(d(e,x))=1 - \frac{FMS(e,x,d(e,x))}{max(FMS(e,x,d(e,x))} $ is used so that good matches render these values small and bounded to the $[0,1]$ range). 

We propose  a cost criteria associated with each $\mathcal{P}_e$, structured as a sum of local costs 
{\small 
 \begin{align}
   {\bf C}({\mathcal{P}_e})= \sum_{x}\lambda \, O(e,x) - \epsilon \, O(e,x)\, O(e,x-1/2) \hspace{0.05in}
    \nonumber \\
    +(1-O(e,x)) FM(e,x,d(e,x))\, , 
    \label{eq:CMF}
\end{align}
}
where the hyperparameters $\lambda, \epsilon$  estimates, for a given path, if the quality of the match between features  at each $(e,x)$ is better than assigning to it an  occlusion $O(e,x)=1$. The lower $\lambda$ is the better the match must be in order not to be chosen as an occlusion. The higher $\epsilon$, the more likely $O(e,x)O(e,x\pm 1/2)=1$.


Thus, the solution $\mathcal{S}^*=(\{[O^*(e,x),d^*(e,x)]; \, x=0,\frac{1}{2}, 1, \frac{3}{2}, \hdots N-1, N-\frac{1}{2}\}$, for each epipolar line $e$, minimizes \eqref{eq:CMF}. 
DP is an optimization method for obtaining the optimal set of disparities $\{d^*(e,x)\}$ and binary variables $\{O^*(e,x)\}$, for each epipolar line $e$. 

Given a state $[O(e,x),d(e,x)]$, and due to the continuity of surfaces and / or occlusions (and then the constraint $GC1$ \eqref{eq:GC1-DP} is used), the DP search is restricted to the following six previous neighbors
{ \begin{align}
    {\cal N}_{[O(e,x),d(e,x)]}\hspace{2.2in}
    \nonumber \\
    =\left \{ [O(e,x- 1/2)=0,1;  d(e,x-1/2)=d(e,x)\pm 0,1/2\right\}\, . \nonumber 
    \label{eq:Neighbors-DP}
\end{align}
}
Note that the pair $[O(e,x)=1,d(e,x)]; [O(e,x-\frac{1}{2})=1,d(e,x-\frac{1}{2})=d(e,x)]$ does not satisfy $GC1$, but is still considered to capture homogeneous regions. 

\subsection{Detection of Homogeneous Regions}
The detection of homogeneous regions—areas lacking texture—occurs during occlusion detection in DP. Occlusions in homogeneous regions arise due to the absence of a good match, as all stereo features in these regions are equally poor matches.
More precisely, we simply assign $h(e,x)=1$ and  $h(e,x-\frac{1}{2})=1$ to  locations $(e,x), (e,x-\frac{1}{2})$ where DP assigned $O(e,x)=O(e,x-\frac{1}{2})=1$ and yet \eqref{eq:GC1-DP} is not satisfied, i.e., where also $0=d(e,x)-d(e,x-\frac{1}{2})$.  
An illustrative example of the results of DP is shown in Figure~\ref{fig:b2fs_workflow} Step (8). Note that the disparity values at the occlusions and homogeneous regions are not reliable values and will still have to be estimated.


We make the DP available on Github\footnote{https://github.com/SherlonAlmeida/B2FS.git}. 

\section{Back to the Future}

\begin{figure*}[!ht]
    \centering
    \includegraphics[width=\textwidth]{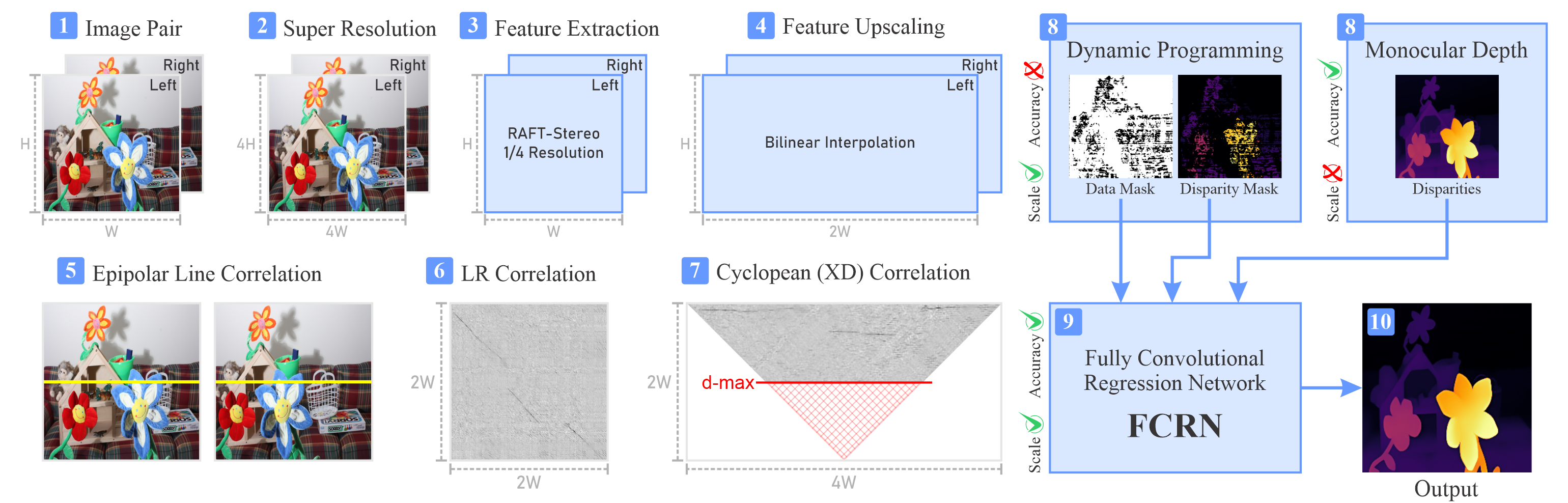}
    \caption{\textbf{Workflow.} Step (1) our approach receives an image-pair with resolution $r^{H,W,3}$. In order to obtain full resolution features and in anticipation of RAFT-Stereo resolution reduction, Step (2)  upscale to $r^{4H,4W,3}$ using Hybrid Attention Transformer~\cite{chen2023activating}. Step  (3), RAFT-Stereo performs  at $\frac{1}{4}$ of $r^{4H,4W,3}$, resulting in $r^{H,W,256}$ for ${F^L, F^R}$. Step (4) uses bilinear interpolation to produce twice as much the width resolution $r^{H,2W,256}$, and achieve the subpixel data for the XD space (the red dots in Figure~\ref{fig:space_transformation}). Steps (5) and (6) perform the feature dot products to create the correlation matrix by epipolar lines at resolution $r^{2W,2W,1}$. Step (7) transfer the  correlation data to the XD space, with the max disparity considered shown by  a solid red horizontal line (cutting the need to search beyond such disparity). Step (8) run DP to obtain a Disparity mask at good data matching coordinates i.e., where a binary Data mask indicates the non occluded and non homogeneous regions. Step (8) also performs monocular depth. Step (9) performs the fill in information of the surface where DP did not have a solution (homogeneous and occlusions) via a FCRN where the input is a normalized monocular depth and the output is the DP disparity map only where such solution is available by the data mask.}
    \label{fig:b2fs_workflow}
\end{figure*}

The model we developed in section~\ref{sec:model} focused on geometric reasoning, a set of abstractions about the 3D scene as viewed from the cyclopean eye. The DP algorithm employed is a method to obtain occlusions and homogeneous regions as well as disparity (depth) in visible regions (where the stereo features ${F^L, F^R}$ do match well) with subpixel accuracy. We still must obtain such features ${F^L, F^R}$, and invoke a surface model to fill in depth at occlusions and homogeneous regions, and possibly improve the accuracy of the depth obtained from DP to a decimal value. We refer to this approach as B2FS  and it is depicted in Figure~\ref{fig:b2fs_workflow} with all steps of B2FS shown and described. We now describe the steps in more depth and at a high level, pointing to the figures and their captions for more detailed information. 



\subsection{Image Features for Stereo Matching}
\begin{figure}[!ht]
    \centering
    \includegraphics[width=0.8\columnwidth]{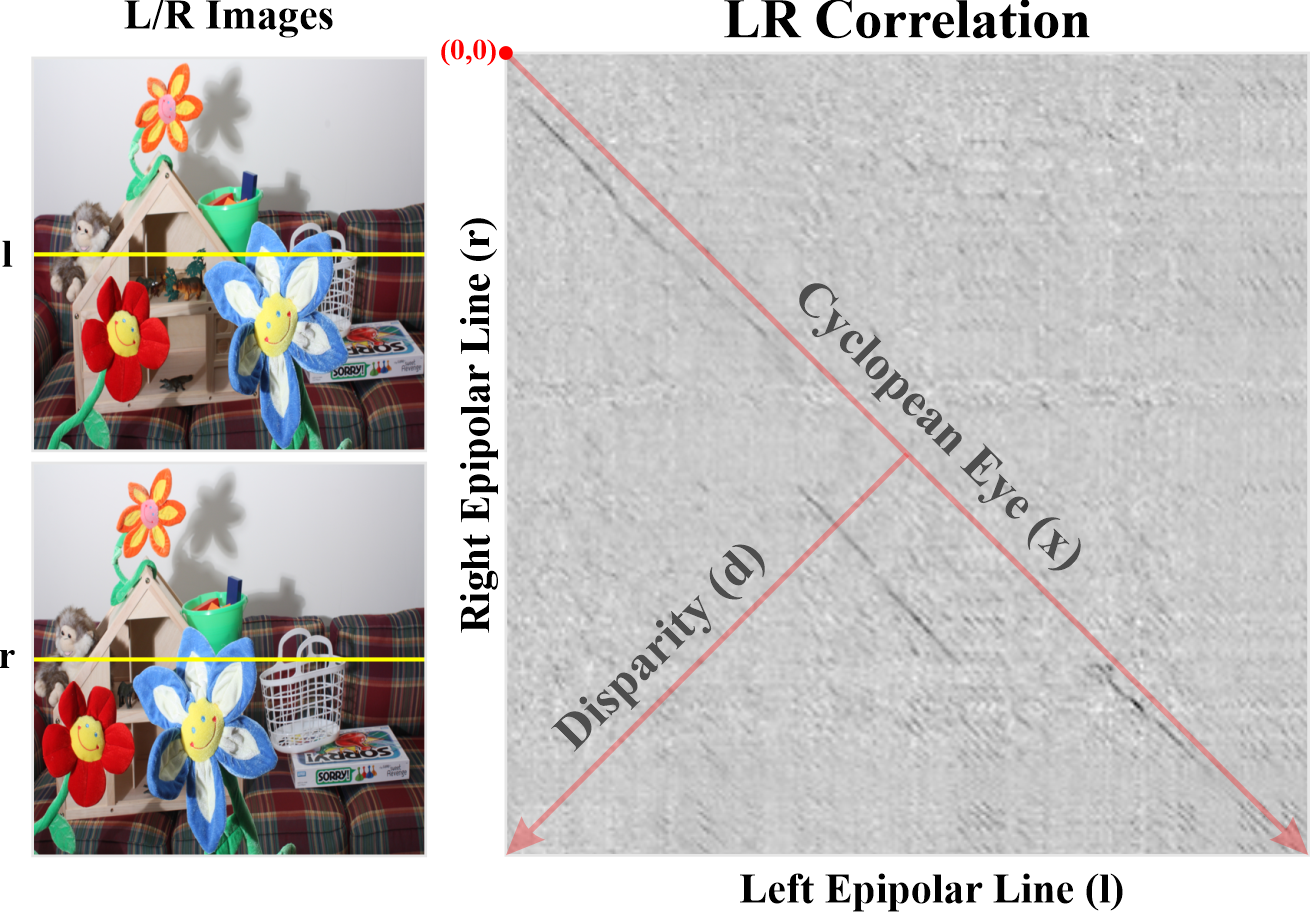}
    \caption{\textbf{LR Correlation:} The graph displays the distance $FM(e,x,d(e,x))=1 - \frac{FMS(e,x,d(e,x))}{max(FMS(e,x,d(e,x))} $, where $FMS$ is obtained from \eqref{eq:fm}. Dark regions (low values of $FM$) represent good matches. The disparity inverse relation to depth guarantees that disparities are always equal or greater than zero, avoiding the search for negative disparities. One observes that typically one disparity stands out (a good match), unless we have occlusions or homogeneous regions. One can identify regions of occlusions where (i) no good match along the disparity axis occur and where (ii) the occlusion width is similar to the disparity changes from neighbor dark segment, as proposed by constraint $GC1$. }
    \label{fig:lr_correlation}
\end{figure}

Considering the promising results from the DL methods already developed that extract stereo features from stereo image pairs, we adopted RAFT-Stereo~\cite{lipson2021raft} features as input to the DP program (see Figure~\ref{fig:lr_correlation}). However, RAFT-Stereo produces features at $\frac{1}{4}$ of the input image resolution. An interpolation must be added to achieve full resolution features and is shown in Figure~\ref{fig:b2fs_workflow}, steps (6) and (7), leading to the LR correlation and data filling of the XD space. Note that Figure~\ref{fig:lr_correlation} already shows the LR correlation at full resolution (after interpolation).

\subsection{Surface Model}



Our approach to a prior model surfaces to complete the depth at occlusions and homogeneous regions, relies on results in monocular depth using DL. Excellent depth gradient information is obtained with Depth Anything V2~\cite{yang2024depthv2} and DepthPro~\cite{bochkovskii2024depth}. As we experimented, the depth values offered are not scaled to precision (see Figure~\ref{fig:quantitative_results_bf2sxdepthpro} for the analysis). Nevertheless, the accuracy of the depth-gradient information, which gives the surface normal unit vectors, is in par with the decimal disparity accuracy we aim to obtain. 


After conducting a study on Depth Anything V2 and DepthPro we observed that the range of the 3D environment captured for DepthPro is wider and edges more accurate than Depth Anything V2. 
Note that monocular depth solutions provide surface information from the view of the provided image and not from the cyclopean eye view. The differences in such depth values are captured by Figure~\ref{fig:monocular_cyclopean_depth} and Equation~\eqref{eq:depth-LR}. 

The challenge is then how to combine both results: the accurate monocular depth gradient or surface normals, from the L/R view,  with the DP output from the cyclopean view. 

\subsection{Combining Stereo x Monocular Depth}

\begin{figure}[!ht]
    \centering
    \includegraphics[width=\columnwidth]{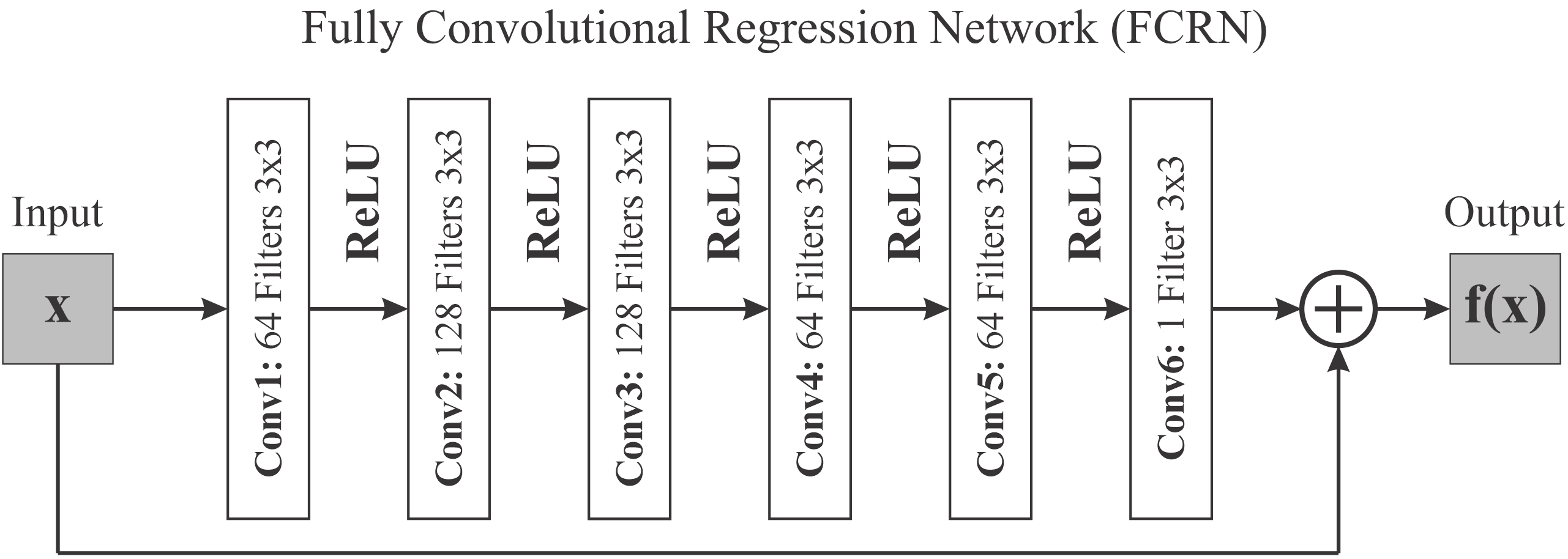}
    \caption{\textbf{FCRN architecture:} The input is the DepthPro disparities, normalized to the range \([0,1]\), and the desired output is the DP subpixel disparity multiplied by the data mask. The regression is conducted using the AdamW optimizer and MSELoss, with an adaptive learning rate initialized at \( lr = 0.00025 \) and adjusted by a scheduler with \( \gamma = 0.90 \) and a step size of 25. The final disparity map is produced after 125 epochs. }
    \label{fig:model_architecture}
\end{figure}

We integrate the surface normal from monocular depth with the output of DP (the disparity and the data mask) via a Fully Convolutional Regression Network (FCRN)
composed of a sequence of six (6) Conv2D layers followed by ReLU activation function (see Figure~\ref{fig:model_architecture}).

The regression is guided using MSELoss, applied only to the data coordinates obtained from DP through stereo matching. Finally, we employ Hybrid Attention Transformer (HAT)~\cite{chen2023activating} to enhance the edge sharpness of the FCRN solution.


\section{Results and Discussion}
\label{sec:results}



We evaluated our approach using the Middlebury dataset, which consists of realistic indoor scenes. Our analysis includes both quantitative and qualitative findings, primarily for images with \( r^{256,256,3} \) resolution. This resolution was chosen to balance quality and practicality—high enough for meaningful results and real applications (e.g., virtual reality) but not so high that thorough experimentation becomes impractical for our limited academic computing resources. Additionally, since most algorithms are designed for high-resolution images, lower resolutions pose a greater challenge for purely data-driven techniques, highlighting the need for the B2FS approach.  

Several comparisons are provided to illustrate B2FS's performance.

\subsection{B2FS vs Monocular Depth}
\begin{figure}[!ht]
    \centering
    \includegraphics[width=1.0\columnwidth]{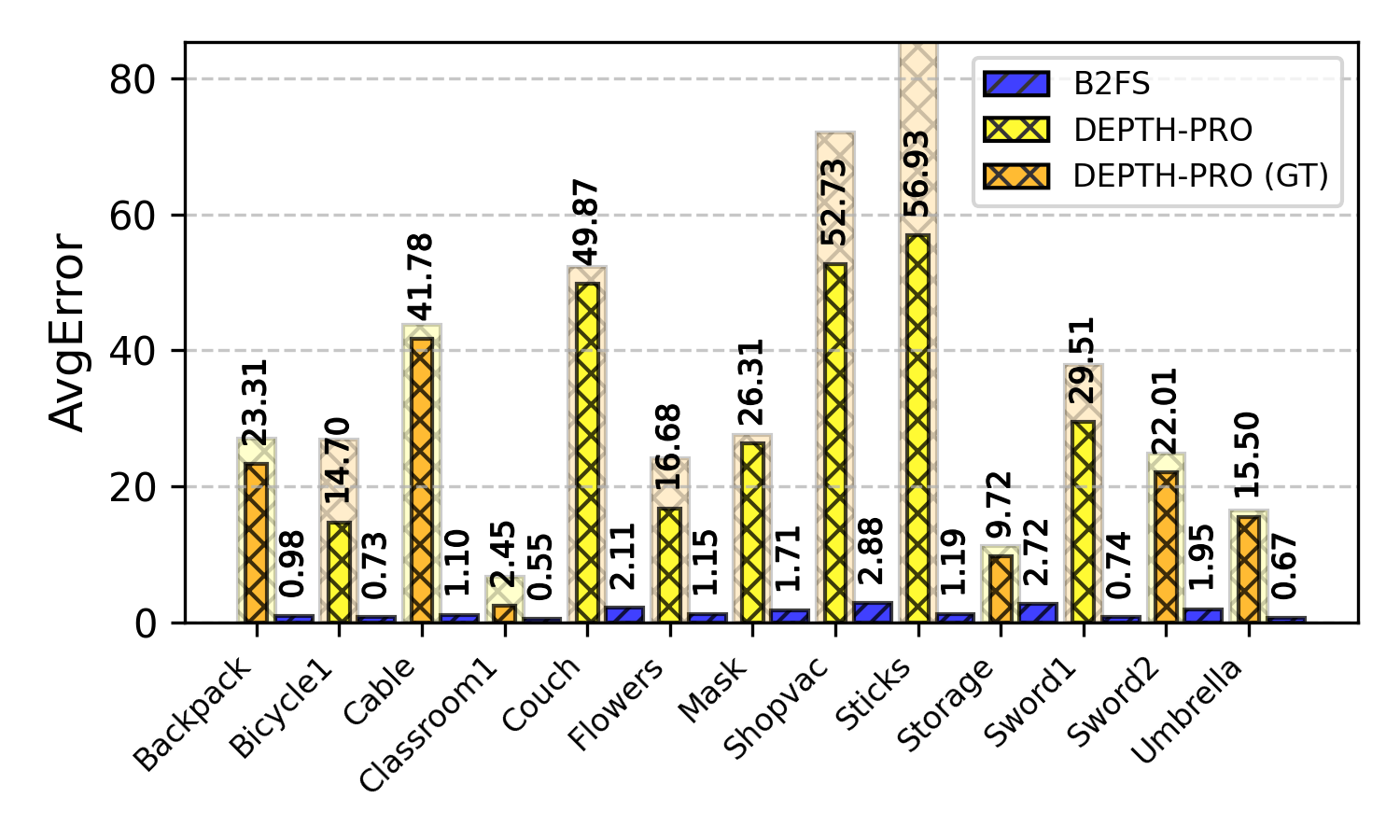}
    \caption{\textbf{Quantitative Results: B2FS vs. DepthPro.} We evaluated the disparity error from Equation~\eqref{eq:disparity-depth}, considering the focal length estimated by DepthPro (yellow) and obtained from GT (orange). In this figure, the best error is represented by the solid-colored bar in the front, while the worst error is overlapped and transparent. Since the focal length estimated by DepthPro was typically better, we adopted it as the default.}
    \label{fig:quantitative_results_bf2sxdepthpro}
\end{figure}

Monocular depth methods "claim to be accurate and they do look good". In order to compare DepthPro depth estimates with the GT disparities, we scale the depth estimates based on the baseline provided by GT and the focal length (either provided by Depth Pro or by the GT). 
Figure~\ref{fig:quantitative_results_bf2sxdepthpro} demonstrate that monocular estimation produces out-of-scale disparities for all images.
In contrast, B2FS obtains accurate solutions across all Middlebury images with public GT (Additional and Training partitions), reducing AvgError by up to 48 times.

\begin{figure}[!ht]
    \centering
    \includegraphics[width=1.0\columnwidth]{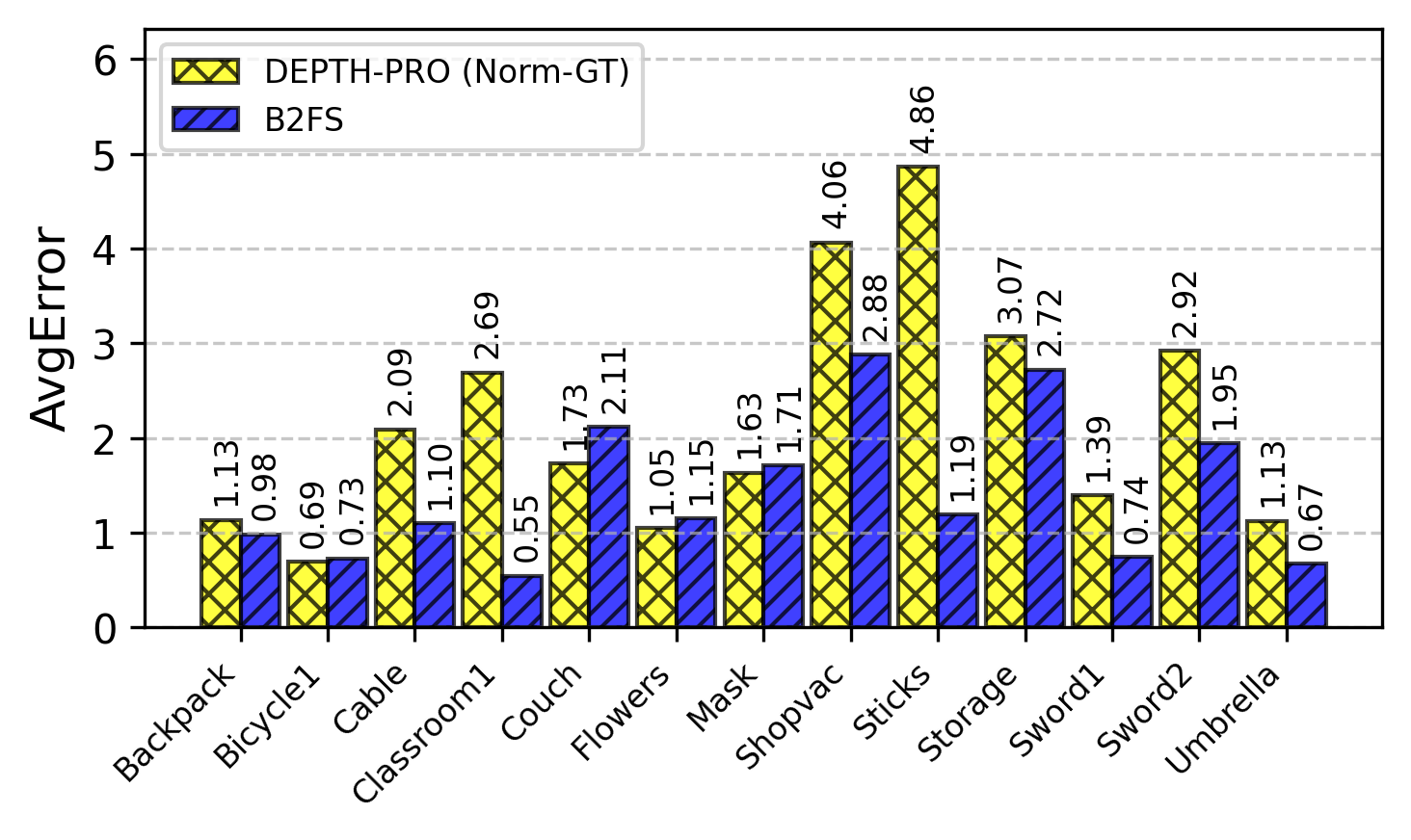}
    \caption{\textbf{Quantitative Results: B2FS vs DepthPro (Normalized-GT).} In this experiment, using GT data,  we applied an affine transformation (translation and scaling) to the DepthPro disparities, guaranteeing it will produce  the minimum and maximum disparities provided by GT. Still B2FS outperform DepthPro, indicating that DepthPro up to an affine transformation can not obtain accurate disparities.}
    \label{fig:quantitative_results_bf2sxdepthpro_norm}
\end{figure}

We then examined the hypothesis of using the best possible affine transformation (translation and scaling) to the monocular depth that would minimize the error with respect to the GT, i.e., we used the GT to estimate the best affine transformation to be applied to monocular depth. Figure~\ref{fig:quantitative_results_bf2sxdepthpro_norm} shows that still B2FS outperforms such result. 


\subsection{B2FS vs RAFT-Stereo, Selective-IGEV}
We obtained the publicly available code for state-of-the-art techniques  RAFT-Stereo~\cite{lipson2021raft} and Selective-IGEV~\cite{wang2024selective} to assess their robustness in extracting stereo cues from input images with lower resolutions than those used in benchmarks such as Middlebury, ETH3D, and KITTI. 

\begin{figure}[!ht]
    \centering
    \includegraphics[width=1.0\columnwidth]{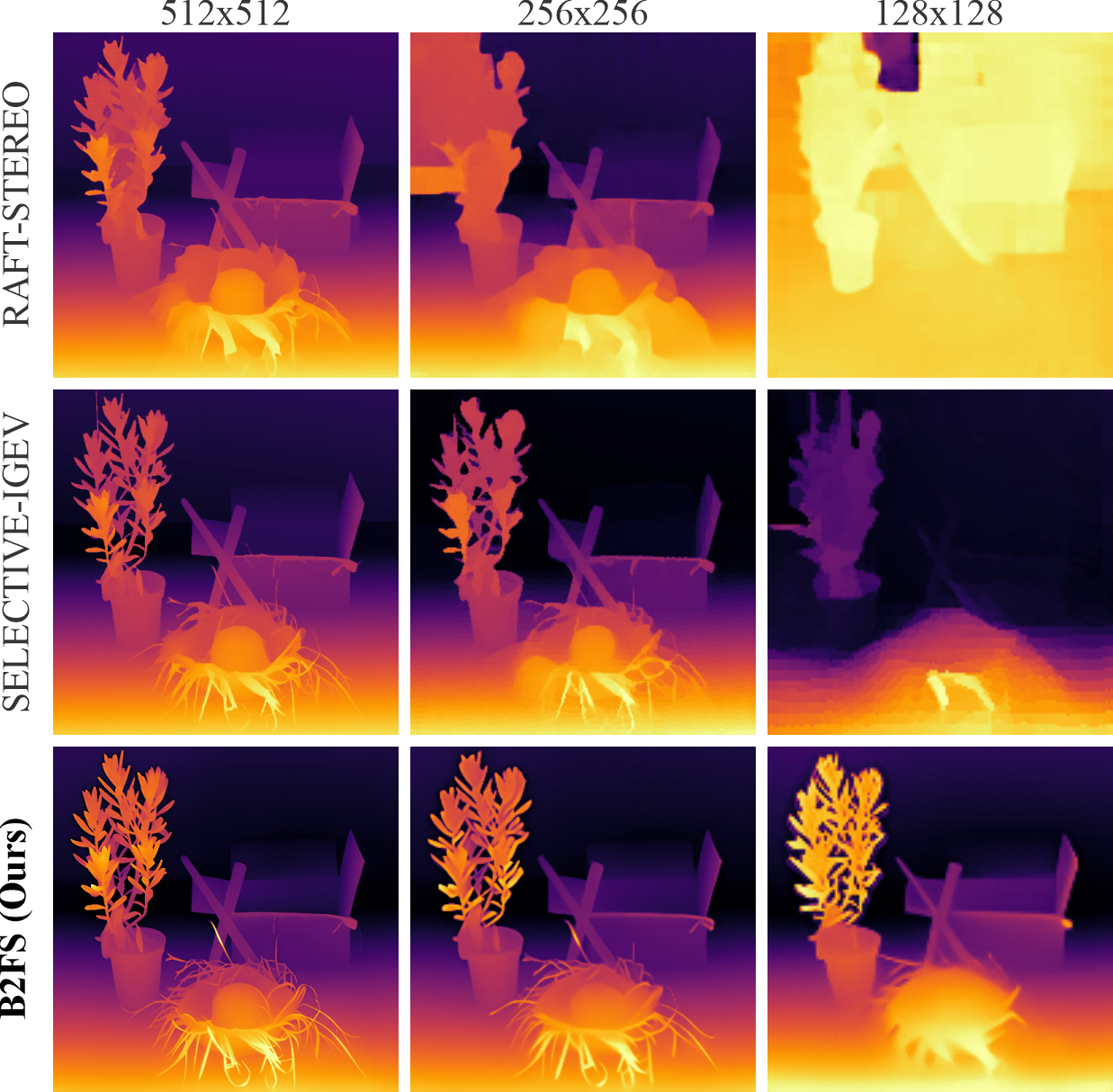}
    \caption{\textbf{Qualitative Results:} A comparison on the impact of input image resolution - more details in Supplementary Material.}
    \label{fig:resolution_impact}
\end{figure}

One study focused on images with resolutions \( r^{512,512,3} \), \( r^{256,256,3} \), and \( r^{128,128,3} \). Figure~\ref{fig:resolution_impact} shows that RAFT-Stereo and Selective-IGEV struggle as the input resolution decreases, while B2FS maintains the surface appearance. Here, we emphasize that B2FS is not overly trained on any particular resolution.

\begin{figure}[!ht]
    \centering
    \includegraphics[width=1.0\columnwidth]{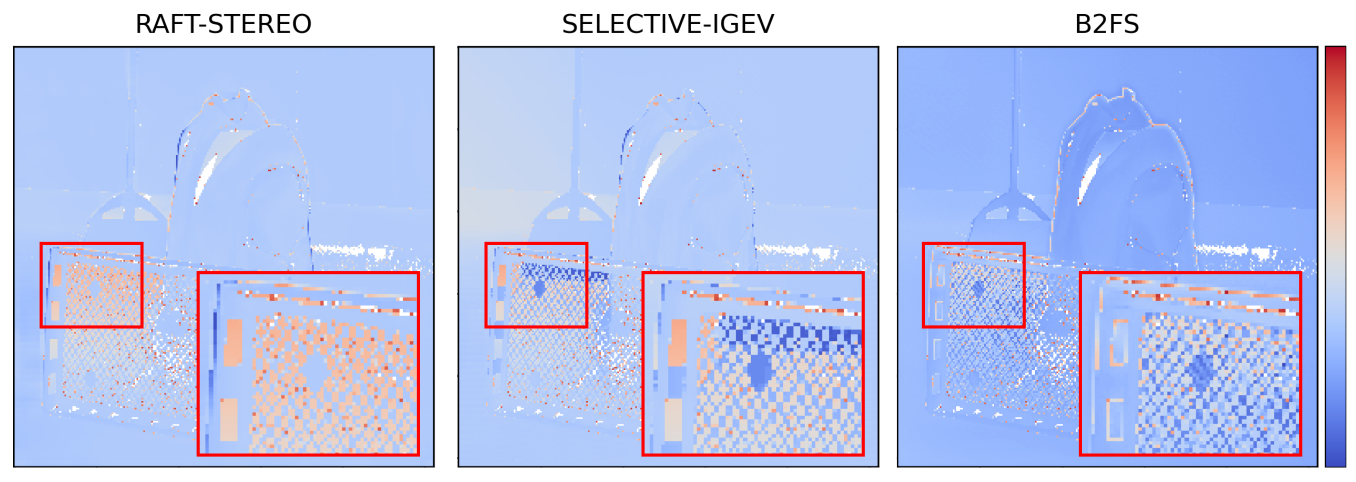}
    \caption{\textbf{Qualitative Results: Signed Error} - red values mean the disparities should be lower (further back), while blue values mean the disparities should be higher (further forward).}
    \label{fig:signed_error}
\end{figure}

\begin{figure}[!ht]
    \centering
    \includegraphics[width=1.0\columnwidth]{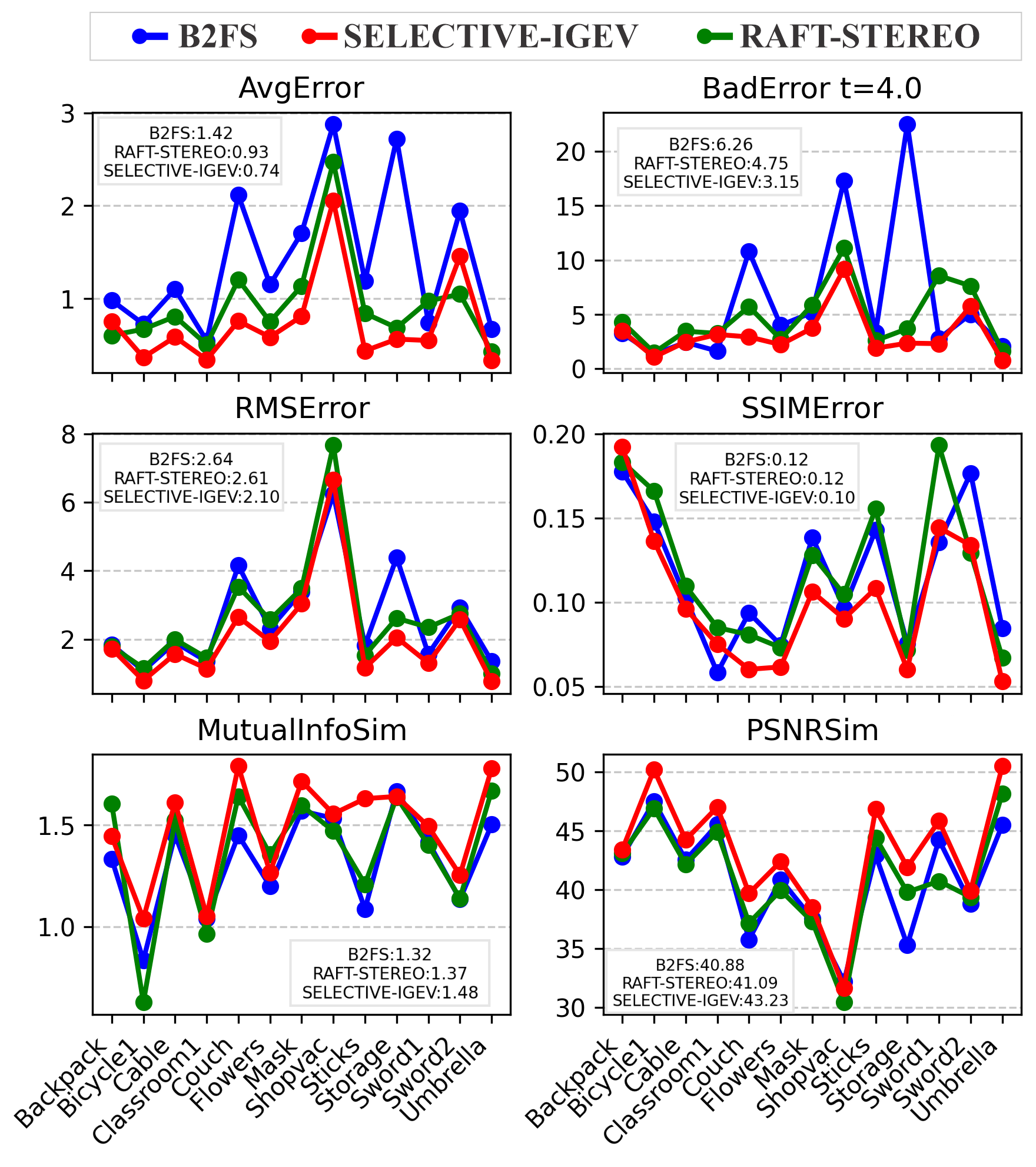}
    \caption{\textbf{Quantitative Results: B2FS vs Stereo Techniques.}}
    \label{fig:quantitative_results}
\end{figure}

Focusing in resolution $r^{256,256,3}$ and standard benchmark with metrics: AvgError, BadError, and RMSError, our numerical results (see Figure~\ref{fig:quantitative_results}) are not superior to RAFT-Stereo and Selective-IGEV.  In close inspection Figure~\ref{fig:signed_error} shows that our approach exhibits small errors everywhere but not large errors in high depth frequency locations relative to RAFT-Stereo and Selective-IGEV, indicating a tendency to our approach to preserve local depth structures.  

Given that our qualitative results preserved detailed surface information while maintaining DP GCs (see Figure~\ref{fig:techniques_comparison}), we questioned the effectiveness of AvgError, BadError, and RMSError in assessing the robustness of depth techniques. To address this, we also evaluated the inverse of the Structural Similarity Index Measure (SSIMError, where lower is better), as well as Mutual Information Similarity (MutualInfoSim) and Peak Signal-to-Noise Ratio (PSNRSim), where higher values indicate better performance (see Figure~\ref{fig:quantitative_results}). Our results (blue) outperformed RAFT-Stereo (green) in most cases and were competitive with Selective-IGEV.

These findings highlight the limitations of traditional error metrics in evaluating depth estimation techniques. While AvgError, BadError, and RMSError provide numerical accuracy assessments, they do not fully capture the perceptual quality or structural consistency of disparity maps. Our results demonstrate that
B2FS preserves finer details and maintains structural coherence, particularly in lower-resolution inputs where stereo-based methods struggle. This suggests that leveraging stereo geometric constraints and monocular gradients can offer a robust alternative, even when input resolutions deviate from those used in training. B2FS effectively balances depth accuracy with high-quality visual reconstruction, making it promising for real-world applications where resolution constraints are common.  

\section{Conclusion}
\label{sec:conclusion}

Back to the Future Cyclopean Stereo (B2FS) addresses a crucial limitation of current AI systems by emphasizing the importance of understanding and abstraction. In an era where models are increasingly data-driven, understanding the process of acquiring knowledge is essential.  

Our proposal neither aims to create a foundation model for surfaces nor dismisses the importance of data-driven models. Instead, we advocate for leveraging pre-trained models as feature extraction and surface prior knowledge about the world and incorporating an additional reasoning step to explore the constraints of specific applications.

Our results highlight the strengths of state-of-the-art monocular depth estimation techniques, while also revealing their limitations in scenarios requiring real-scale depth. Furthermore, we reaffirm that stereo information remains the most reliable source for 3D scene perception, both for AI systems and human vision. By integrating DL with analytical models, we unlock new possibilities for computer vision, not only benefiting from DL's capabilities but also using it to deepen our understanding and interpretation of Stereo Vision and Human Perception.  Note that our approach, in principle,  could be extended straightforwardly to transparent surfaces, relaxing on $GC2$ to allow multiple disparities.

This work paves the way for future research on hybrid AI approaches that balance data-driven learning with structured reasoning. Expanding beyond Stereo Vision, this methodology could enhance AI applications in robotics, medical imaging, virtual reality, and autonomous navigation, where understanding the underlying geometry of the environment is as crucial as pattern recognition.
\section*{Acknowledgements}

This study was financed in part by the Coordenação de Aperfeiçoamento de Pessoal de Nível Superior – Brasil (CAPES) – Finance Code 001.

{
    \small
    \bibliographystyle{ieeenat_fullname}
    \bibliography{main}
}

\clearpage
\maketitlesupplementary

\section{Overview}

In this document, we extend the demonstration of the main contributions of the B2FS approach, emphasizing a visual presentation.  

In Section~\ref{sec:matching-occlusions-homogeneity} we revisit the two proposed geometric constraints $GC1$ and $GC2$ from direct data visualization as well as an inspection of occlusions, homogeneous regions, and data matching.

In Section~\ref{sec:visual_analysis} we provide visualization for the qualitative comparison of maintaining details between RAFT-Stereo, Selective-IGEV, and B2FS for all the 13 images of the additional partition of Middlebury dataset: Backpack, Bicycle, Cable, Classroom1, Couch, Flowers, Mask, Shopvac, Sticks, Storage, Sword1, Sword2, and Umbrella.

In Section~\ref{sec:signed_error} a visual comparison of the disparity errors per pixel suggests that B2FS performs significantly better at depth discontinuities, but slightly underperformed at homogeneous regions due to B2FS discarding feature matching there. Since homogeneous regions are large regions of the image, B2FS end up with more total error. Future work on integrating feature matching in these regions could improve the results. All images presented in Section~\ref{sec:visual_analysis} and Section~\ref{sec:signed_error} have a resolution of $r^{256,256,3}$ and include a caption that specifies the image of the data set used.

Finally, we present in Section~\ref{sec:image_resolution} a study about the impact on the results of changing image resolution, and Section~\ref{sec:general-comments} concludes this document with general comments.

\section{Matching, Occlusions and Homogeneity}
\label{sec:matching-occlusions-homogeneity}
\begin{figure*}[!ht]
    \centering
    \includegraphics[width=0.90\textwidth]{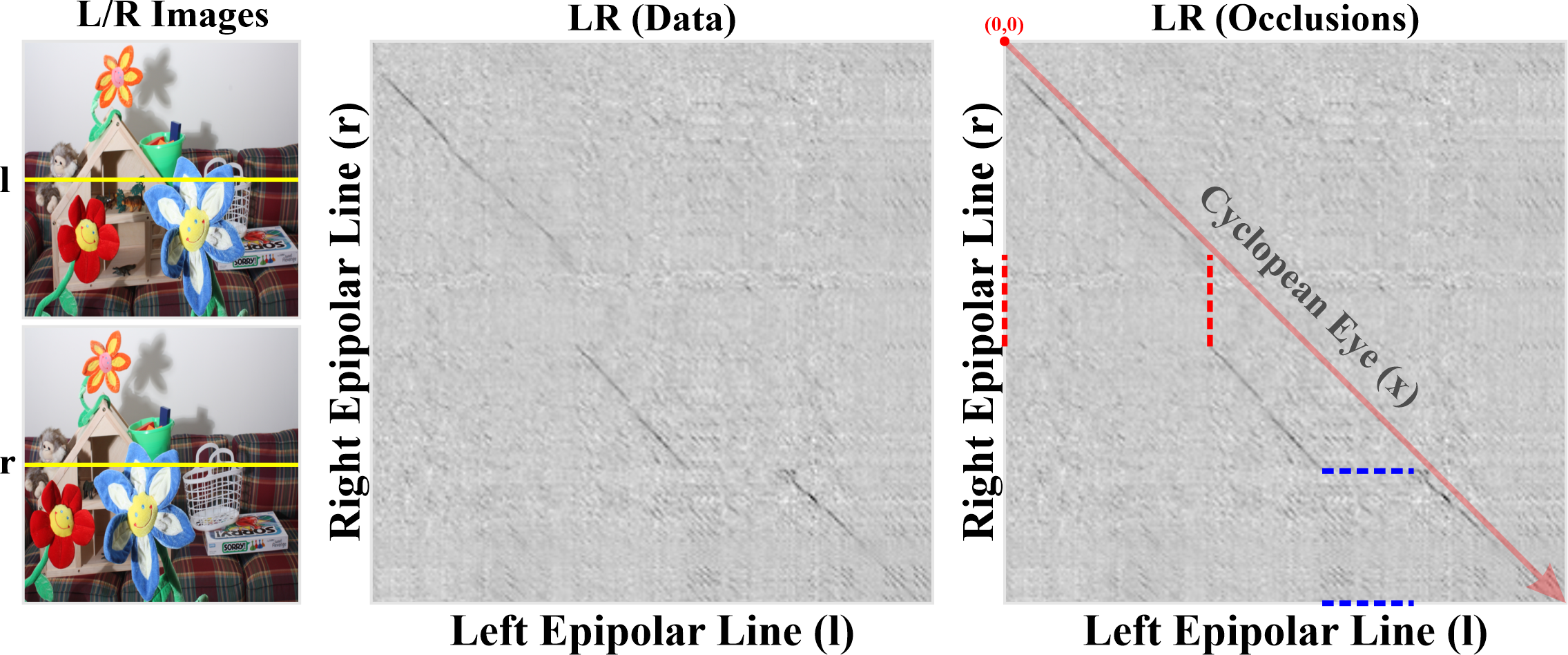}
    \caption{The LR space displays $FM$ distances associated with the epipolar lines $e=128$ (yellow), where $FM_{e,x}(d) \in [0,1]$. Dark regions represent  low $FM$ and good matches. $GC2$ is verified with R-occlusions in dashed-red, no matches, associated with the L-discontinuity (vertical jumps) and   L-occlusions in dashed-blued, no matches, associated with R-discontinuity (horizontal jumps). From the cyclopean eye these are 45 degrees changes and $GC1$, unique disparity,  is also verified.}
    \label{fig:occlusions}
\end{figure*}

\begin{figure*}[!ht]
    \centering
    \includegraphics[width=0.90\textwidth]{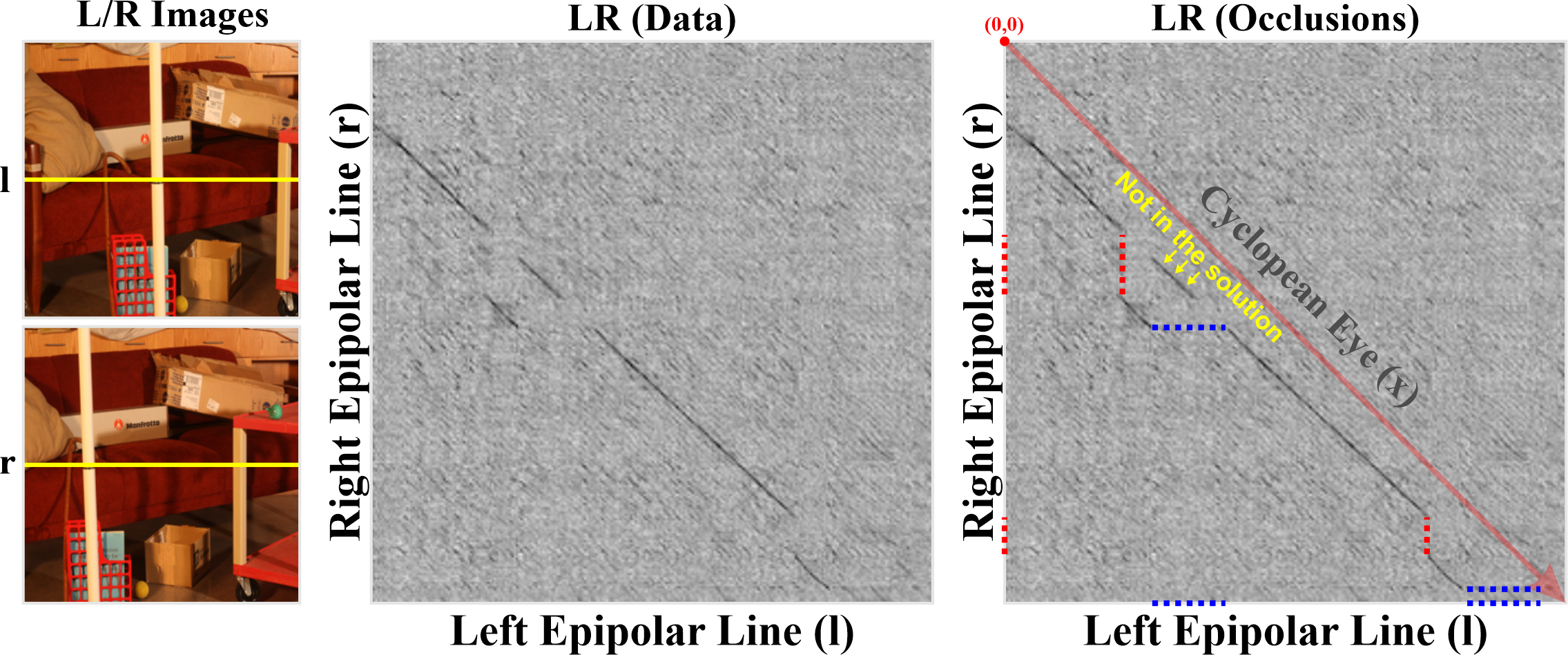}
    \caption{The LR space displays $FM$ distances associated with the epipolar lines $e=128$ (yellow), where $FM_{e,x}(d) \in [0,1]$. Dark regions represent  low $FM$ and good matches. $GC2$ is verified with R-occlusions in dashed-red, no matches, associated with the L-discontinuity (vertical jumps) and   L-occlusions in dashed-blued, no matches, associated with R-discontinuity (horizontal jumps). From the cyclopean eye these are 45 degrees changes. In particular in order for stereo to output the front post depth (disparity)  $GC1$, a unique disparity constraint, will have to be imposed to discard other possible good feature matches (low $FM$ 
 values) as shown in the right figure.}
    \label{fig:uniqueness}
\end{figure*}

\begin{figure*}[!ht]
    \centering
    \includegraphics[width=0.70\textwidth]{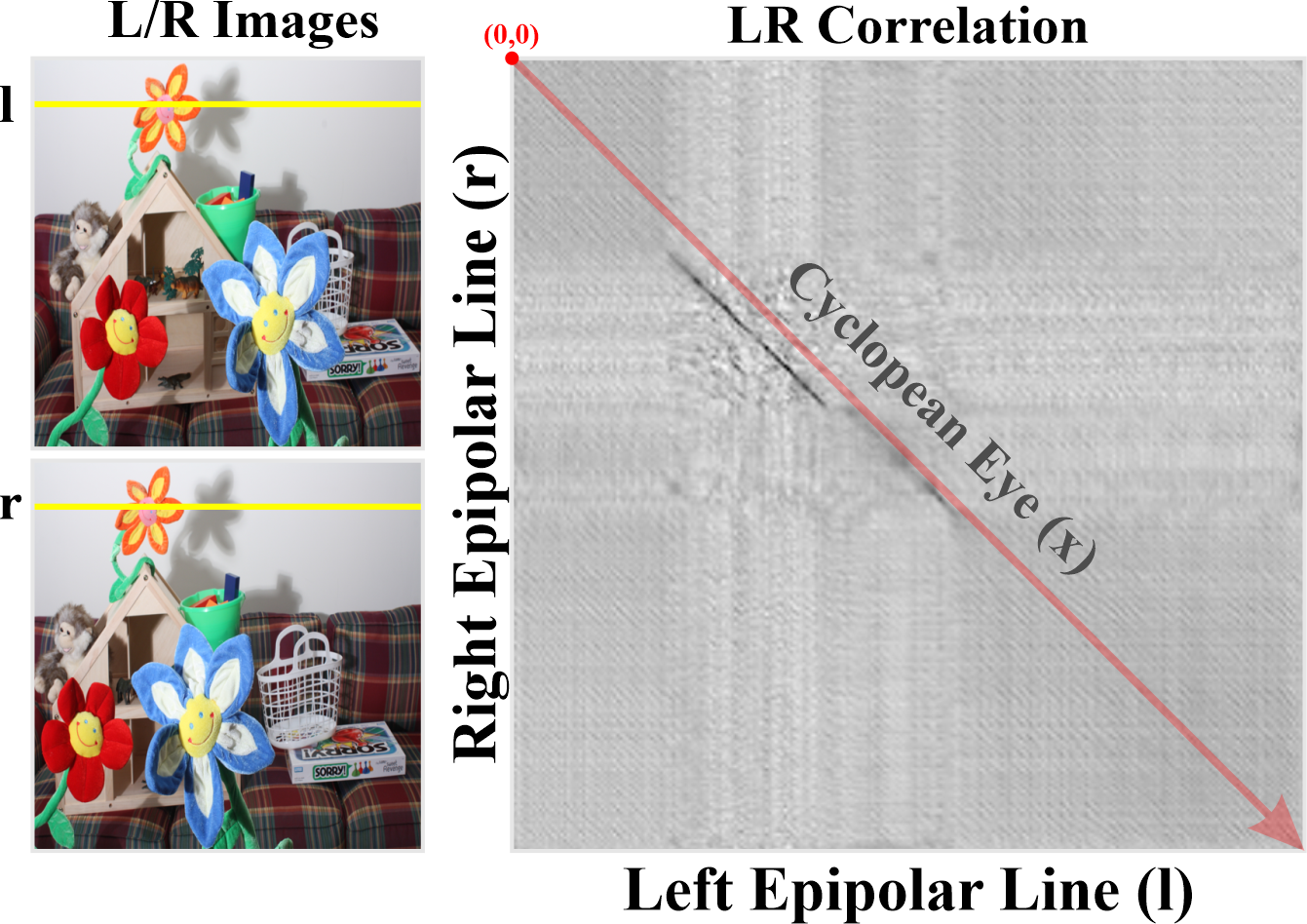}
    \caption{Homogeneity: The LR space displays $FM$ distances associated with the epipolar lines $e=30$ (yellow), where $FM_{e,x}(d) \in [0,1]$. Dark regions represent  low $FM$ and good matches.  Note that shadows do yield good matches. However, in order to fill in homogeneous regions, surface priors become necessary. In our approach, we detect such homogeneous regions and discard feature matching values at these regions. }
    \label{fig:homogeneity}
\end{figure*}

This section examines data to validate the geometric reasoning presented by Proposition 2.4, constraints $GC1$ and $GC2$, and to validate the challenge of feature matching in homogeneous regions. See Figures~\ref{fig:occlusions},~\ref{fig:uniqueness}, and~\ref{fig:homogeneity} and their respective captions.

\section{Qualitative results: a visual analysis}
\label{sec:visual_analysis}
\begin{figure*}[!ht]
    \centering
    \includegraphics[width=1.00\textwidth]{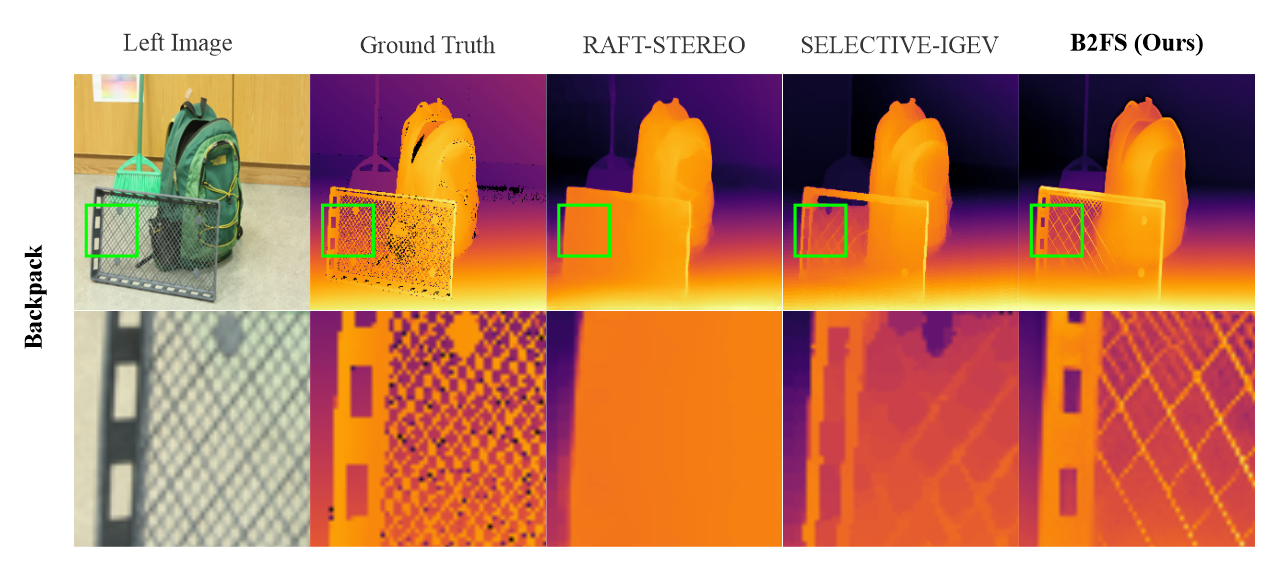}
    \caption{Backpack}
    \label{fig:Backpack_vis}
\end{figure*}

\begin{figure*}[!ht]
    \centering
    \includegraphics[width=1.00\textwidth]{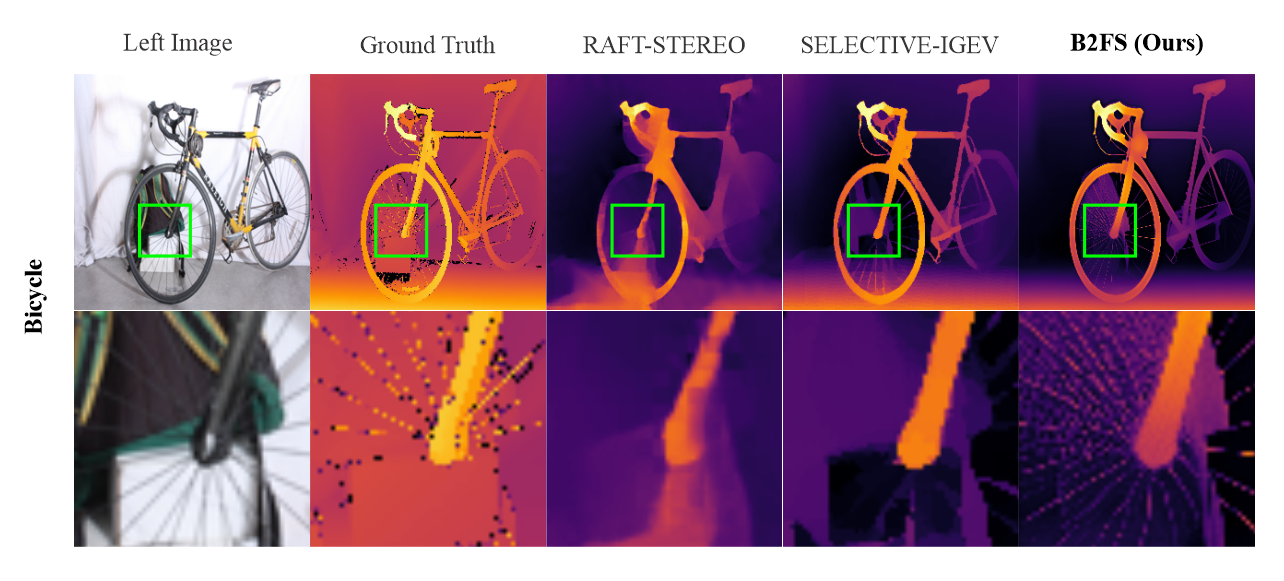}
    \caption{Bicycle}
    \label{fig:Bicycle_vis}
\end{figure*}

\begin{figure*}[!ht]
    \centering
    \includegraphics[width=1.00\textwidth]{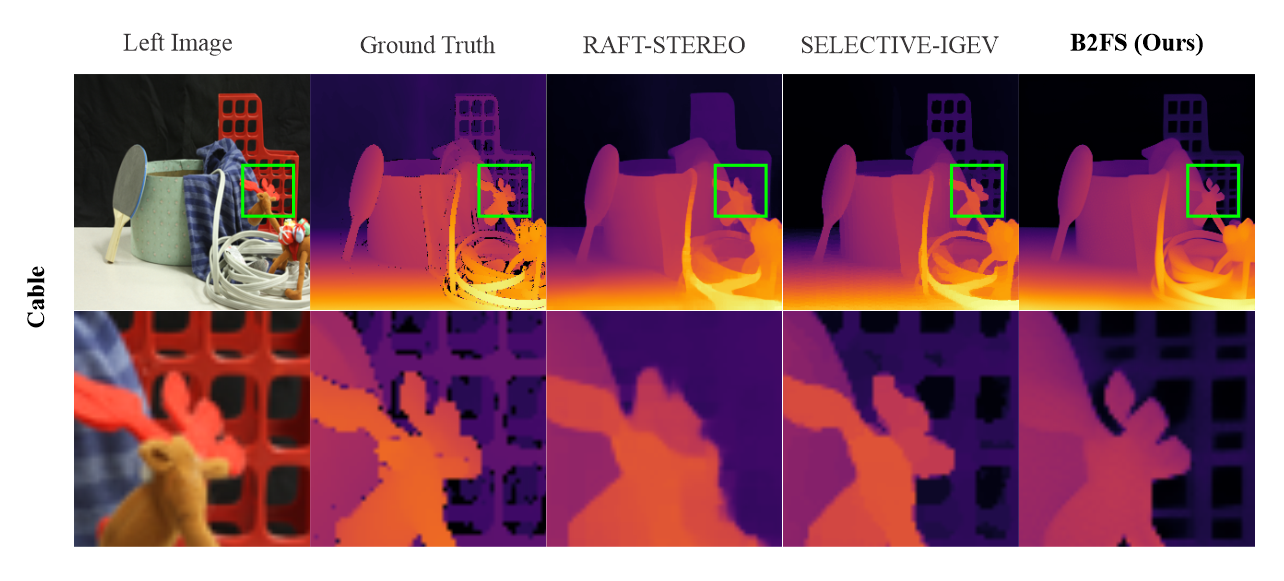}
    \caption{Cable}
    \label{fig:Cable_vis}
\end{figure*}

\begin{figure*}[!ht]
    \centering
    \includegraphics[width=1.00\textwidth]{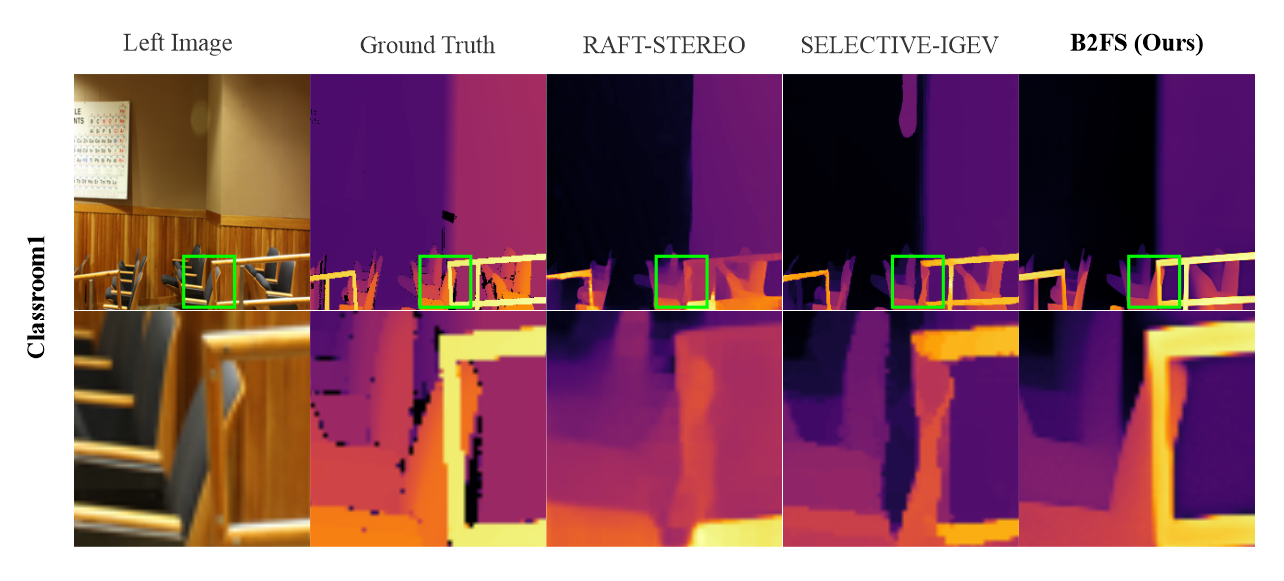}
    \caption{Classroom1}
    \label{fig:Classroom1_vis}
\end{figure*}

\begin{figure*}[!ht]
    \centering
    \includegraphics[width=1.00\textwidth]{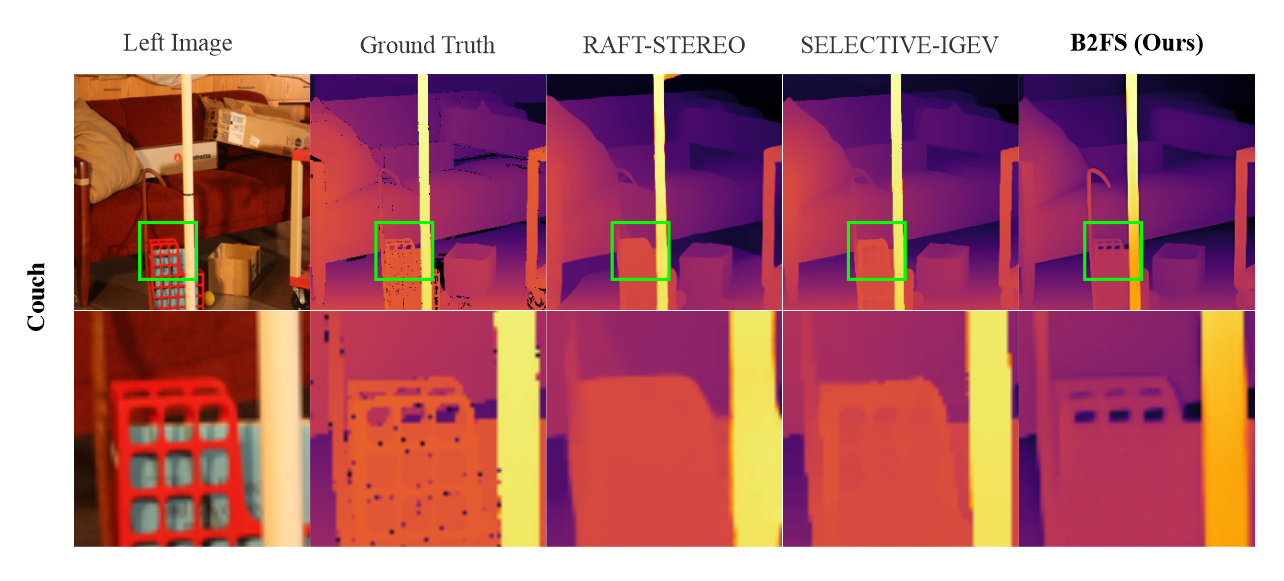}
    \caption{Couch}
    \label{fig:Couch_vis}
\end{figure*}

\begin{figure*}[!ht]
    \centering
    \includegraphics[width=1.00\textwidth]{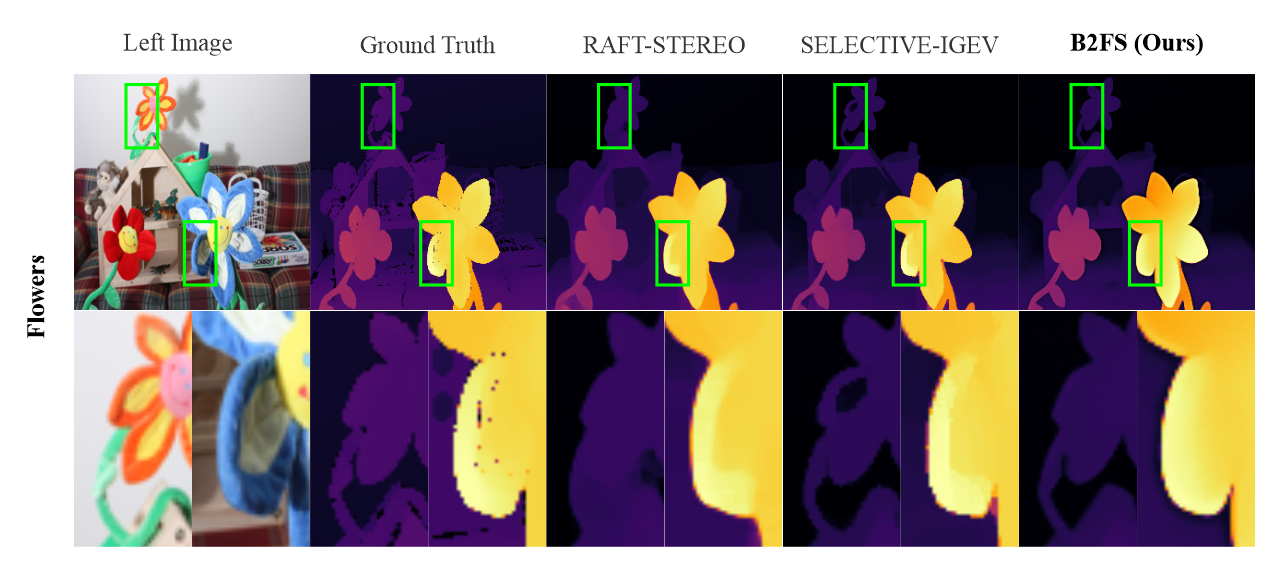}
    \caption{Flowers}
    \label{fig:Flowers_vis}
\end{figure*}

\begin{figure*}[!ht]
    \centering
    \includegraphics[width=1.00\textwidth]{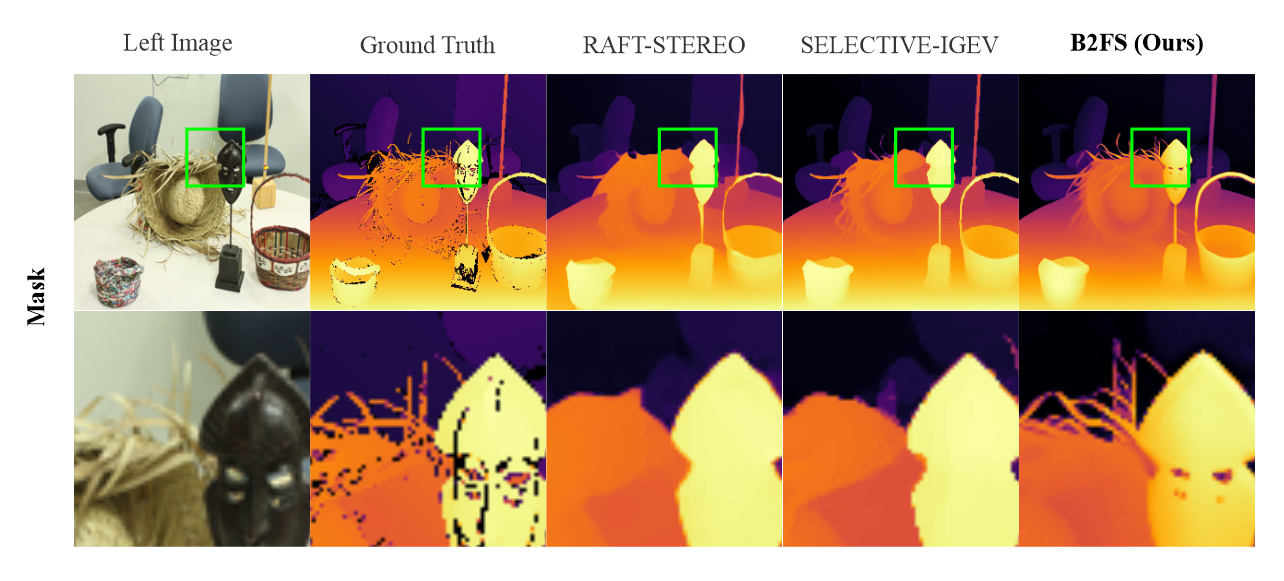}
    \caption{Mask}
    \label{fig:Mask_vis}
\end{figure*}

\begin{figure*}[!ht]
    \centering
    \includegraphics[width=1.00\textwidth]{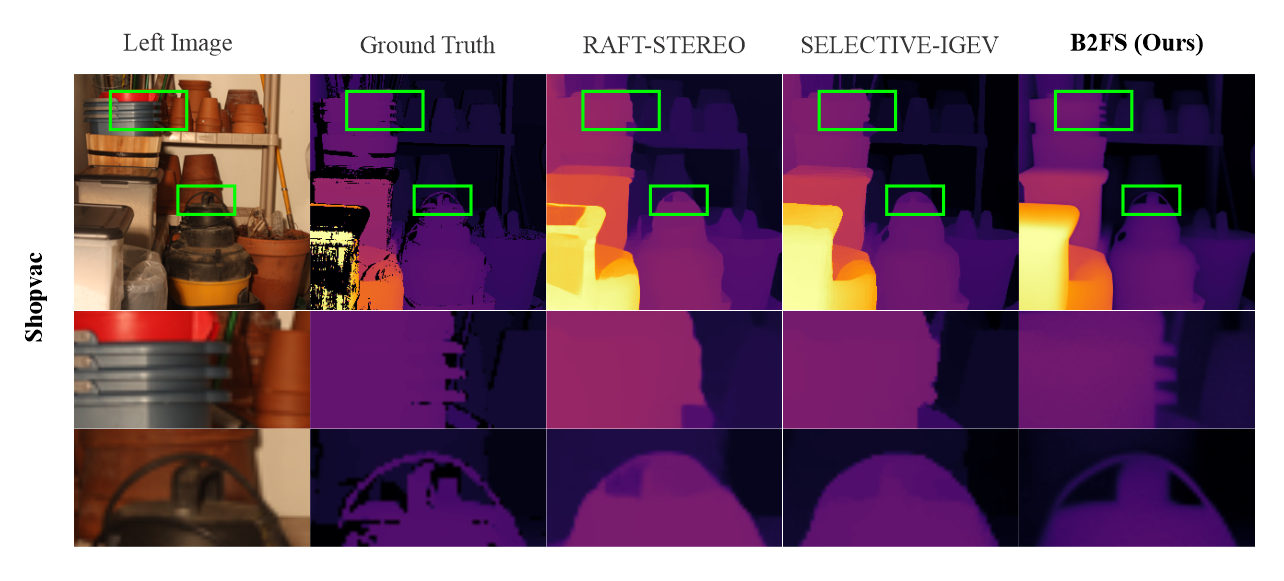}
    \caption{Shopvac}
    \label{fig:Shopvac_vis}
\end{figure*}

\begin{figure*}[!ht]
    \centering
    \includegraphics[width=1.00\textwidth]{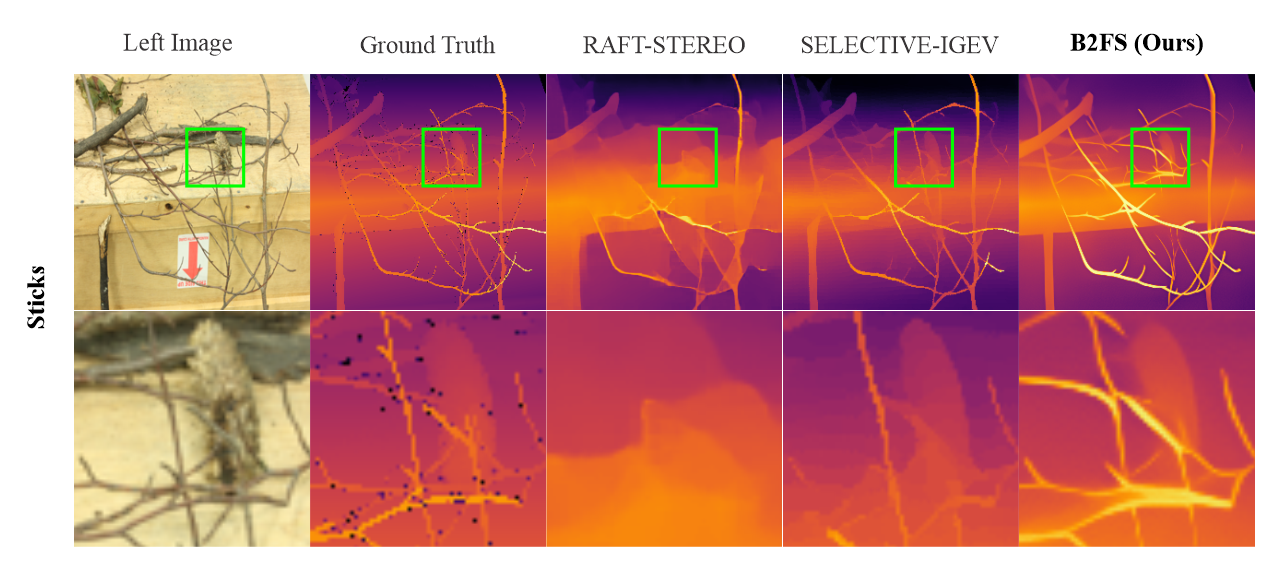}
    \caption{Sticks}
    \label{fig:Sticks_vis}
\end{figure*}

\begin{figure*}[!ht]
    \centering
    \includegraphics[width=1.00\textwidth]{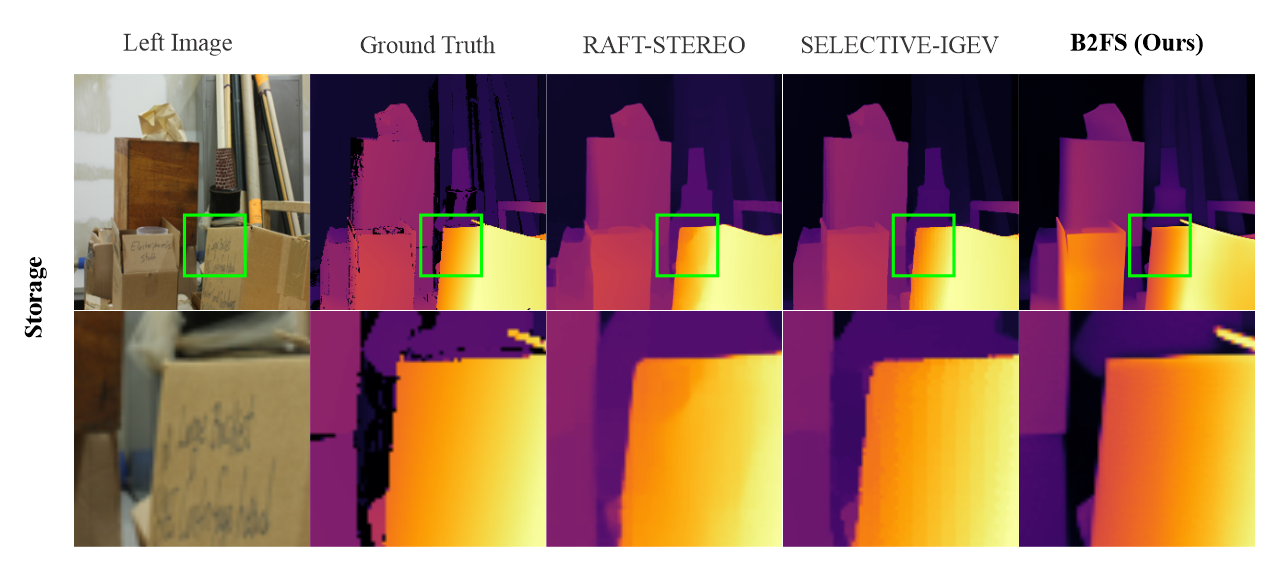}
    \caption{Storage}
    \label{fig:Storage_vis}
\end{figure*}

\begin{figure*}[!ht]
    \centering
    \includegraphics[width=1.00\textwidth]{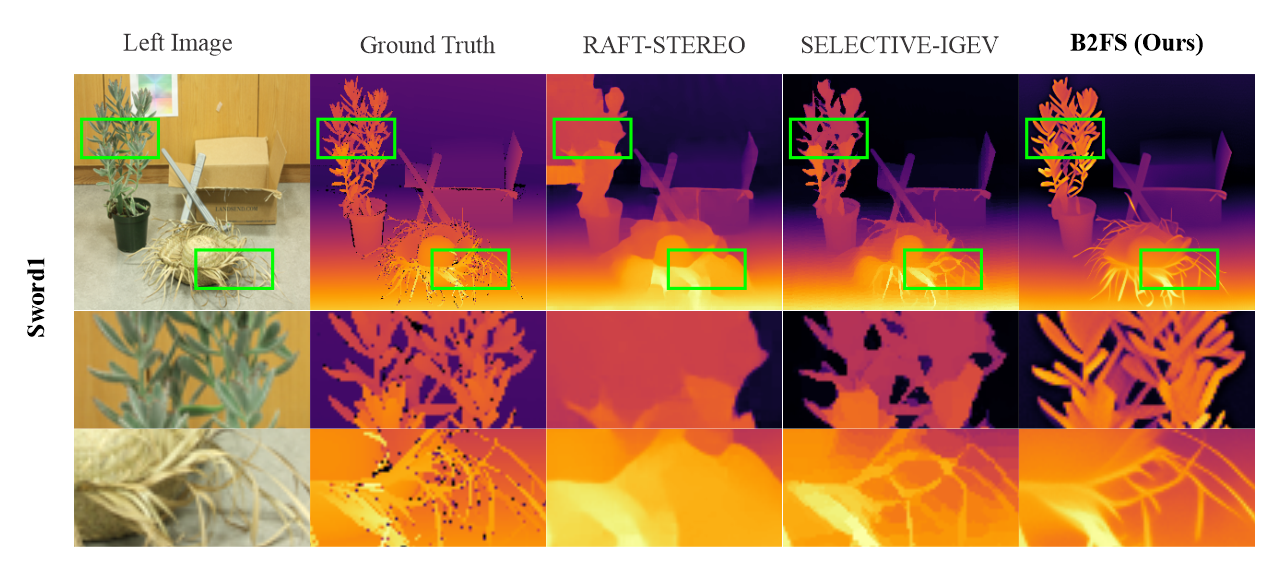}
    \caption{Sword1}
    \label{fig:Sword1_vis}
\end{figure*}

\begin{figure*}[!ht]
    \centering
    \includegraphics[width=1.00\textwidth]{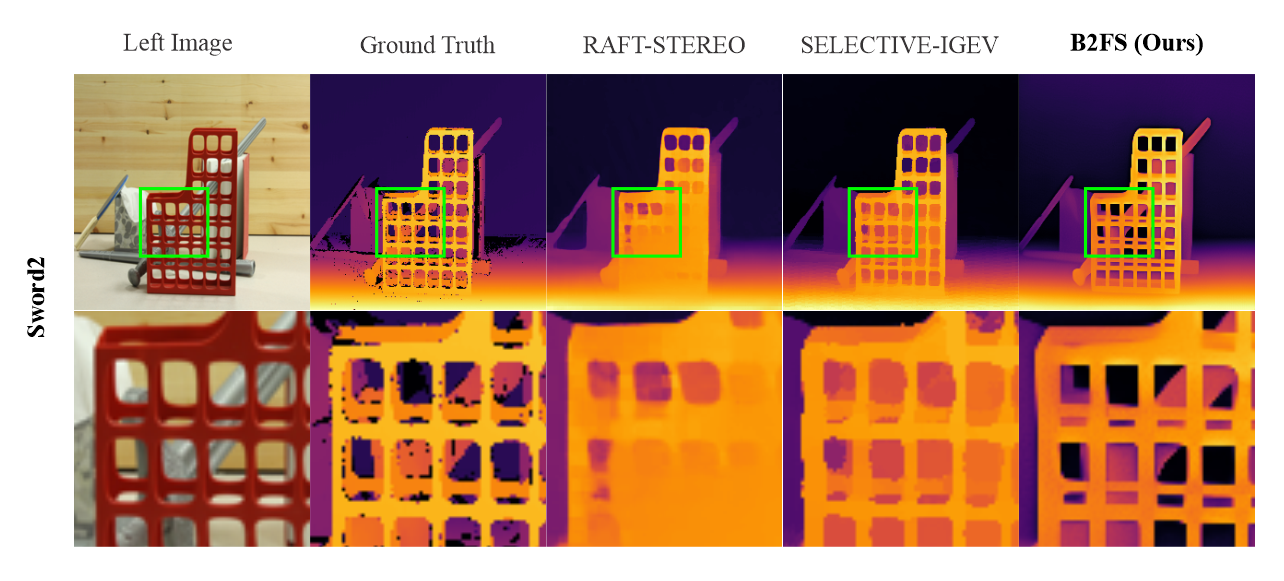}
    \caption{Sword2}
    \label{fig:Sword2_vis}
\end{figure*}

\begin{figure*}[!ht]
    \centering
    \includegraphics[width=1.00\textwidth]{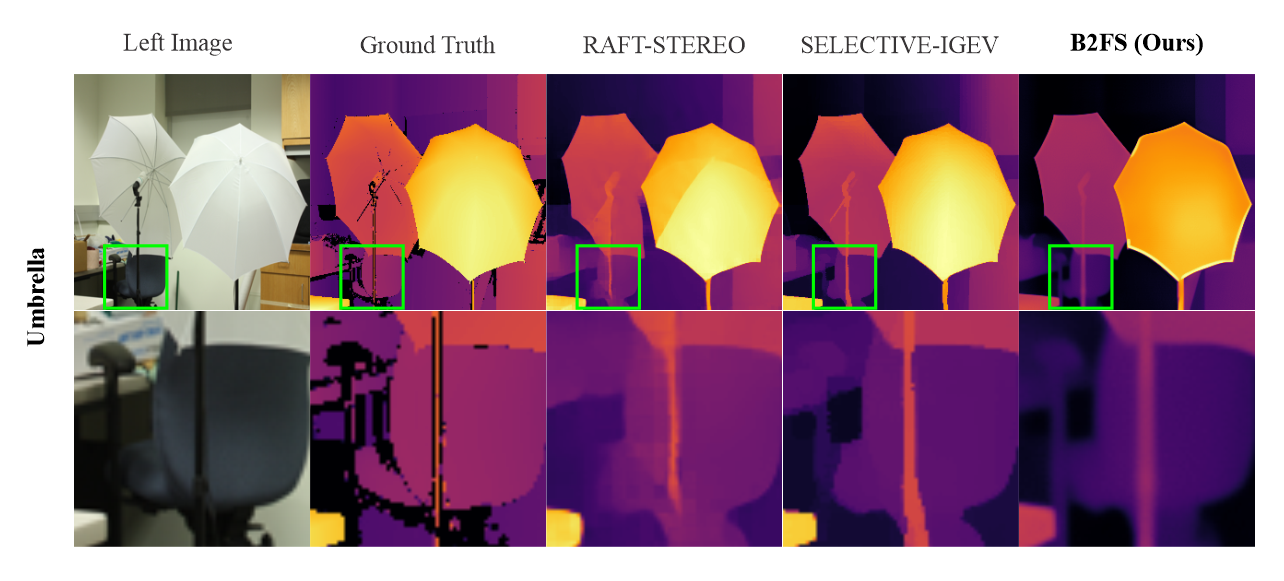}
    \caption{Umbrella}
    \label{fig:Umbrella_vis}
\end{figure*}

For each image, we selected a green rectangle to highlight a region of interest and zoomed in for better visualization. The results support the robustness of B2FS in preserving the scene structure. See figures~\ref{fig:Backpack_vis},~\ref{fig:Bicycle_vis},~\ref{fig:Cable_vis},~\ref{fig:Classroom1_vis},~\ref{fig:Couch_vis},~\ref{fig:Flowers_vis},~\ref{fig:Mask_vis},~\ref{fig:Shopvac_vis},~\ref{fig:Sticks_vis},~\ref{fig:Storage_vis},~\ref{fig:Sword1_vis},~\ref{fig:Sword2_vis},~\ref{fig:Umbrella_vis}.

\section{Quantitative visual results: signed error}
\label{sec:signed_error}

\begin{figure*}[!ht]
    \centering
    \includegraphics[width=1.0\textwidth]{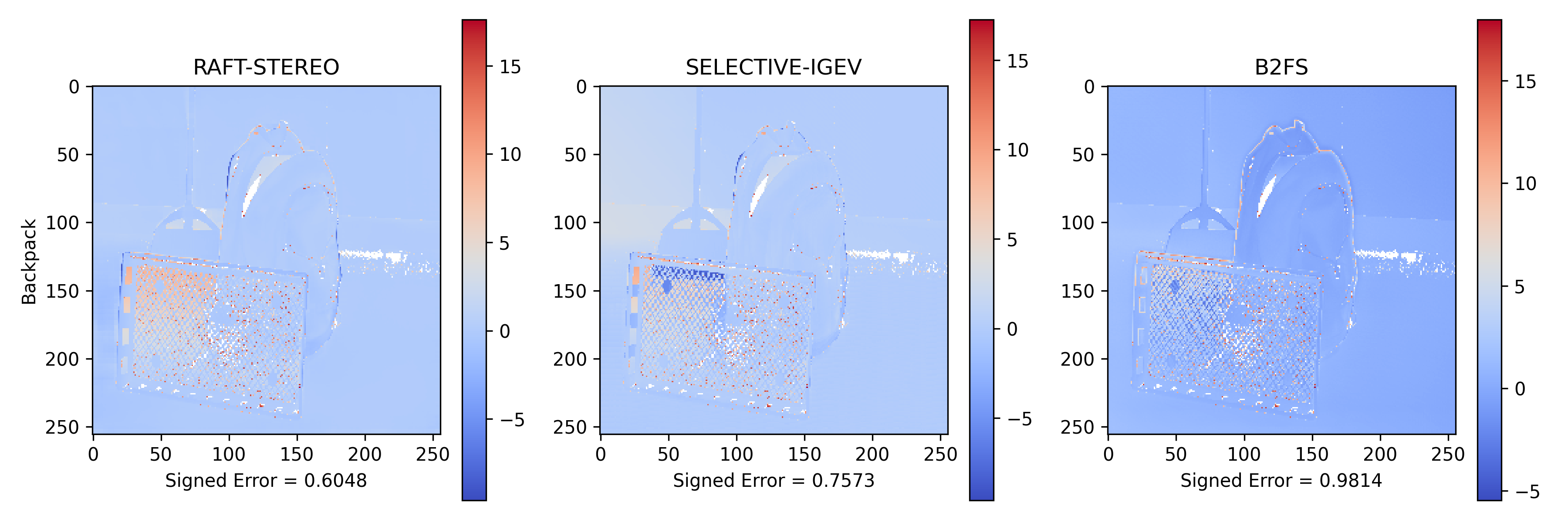}
    \caption{Backpack}
    \label{fig:Backpack_err}
\end{figure*}

\begin{figure*}[!ht]
    \centering
    \includegraphics[width=1.0\textwidth]{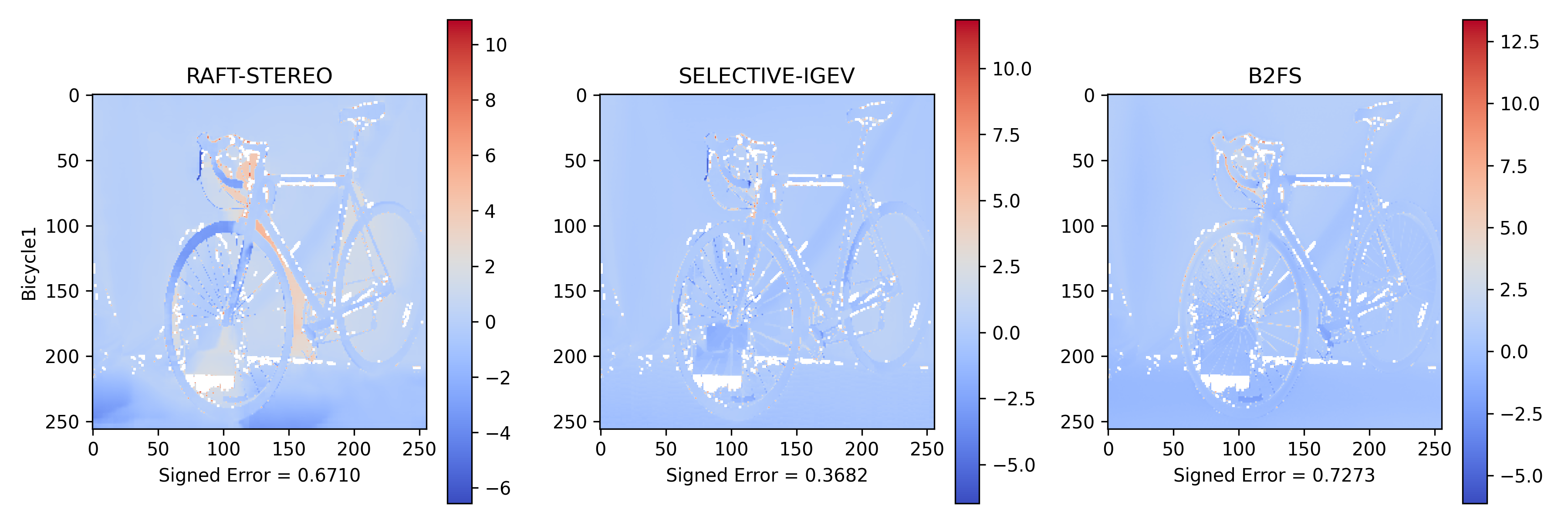}
    \caption{Bicycle}
    \label{fig:Bicycle_err}
\end{figure*}

\begin{figure*}[!ht]
    \centering
    \includegraphics[width=1.0\textwidth]{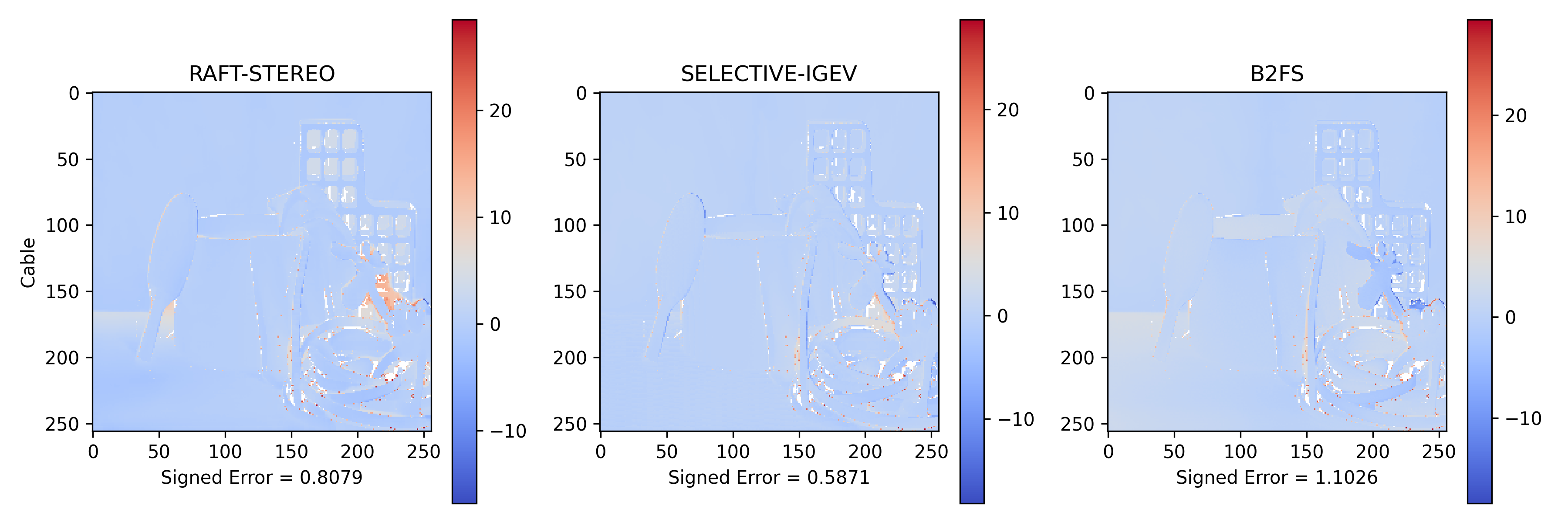}
    \caption{Cable}
    \label{fig:Cable_err}
\end{figure*}

\begin{figure*}[!ht]
    \centering
    \includegraphics[width=1.0\textwidth]{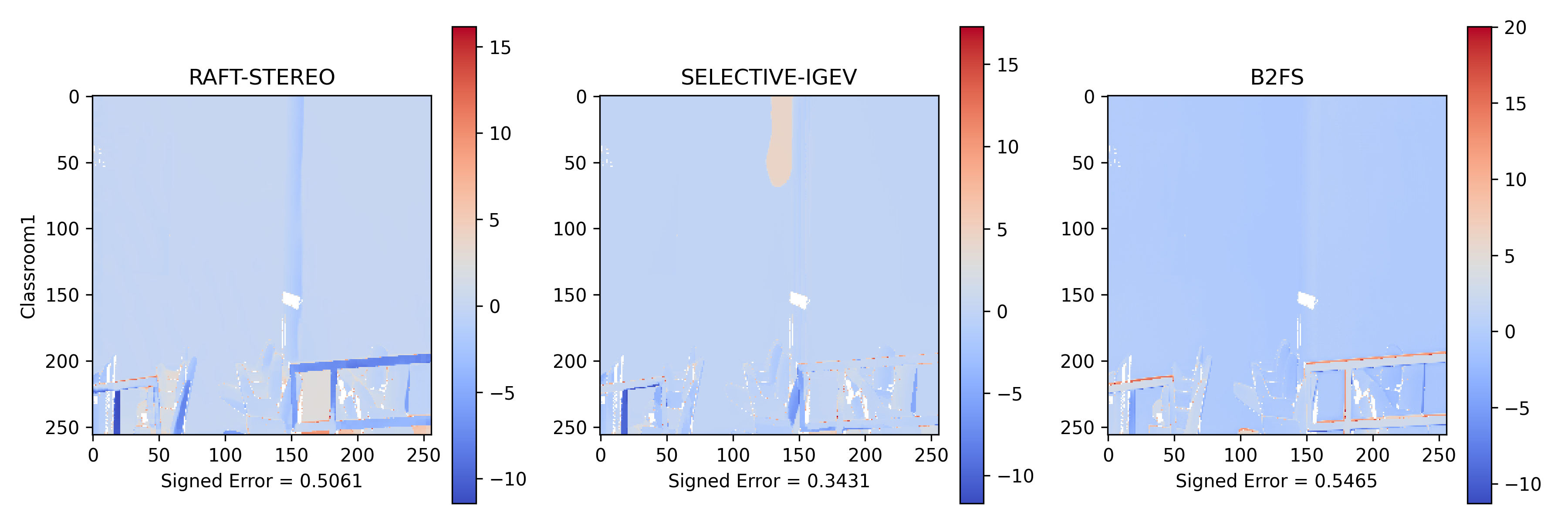}
    \caption{Classroom1}
    \label{fig:Classroom1_err}
\end{figure*}

\begin{figure*}[!ht]
    \centering
    \includegraphics[width=1.0\textwidth]{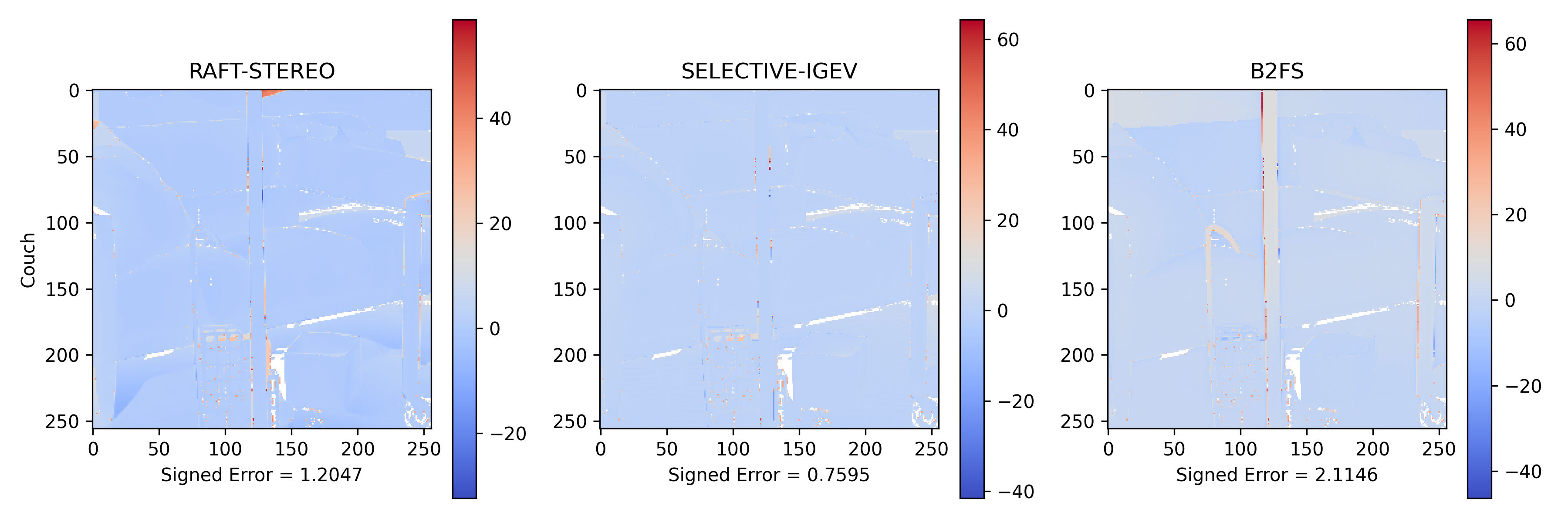}
    \caption{Couch}
    \label{fig:Couch_err}
\end{figure*}

\begin{figure*}[!ht]
    \centering
    \includegraphics[width=1.0\textwidth]{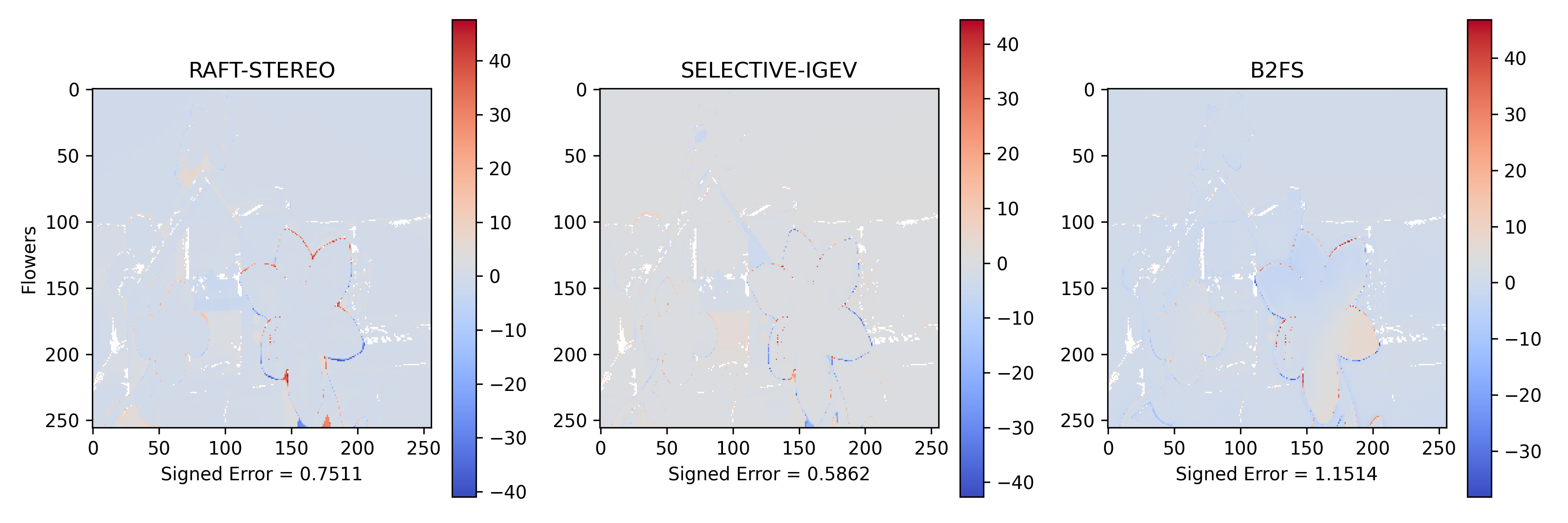}
    \caption{Flowers}
    \label{fig:Flowers_err}
\end{figure*}

\begin{figure*}[!ht]
    \centering
    \includegraphics[width=1.0\textwidth]{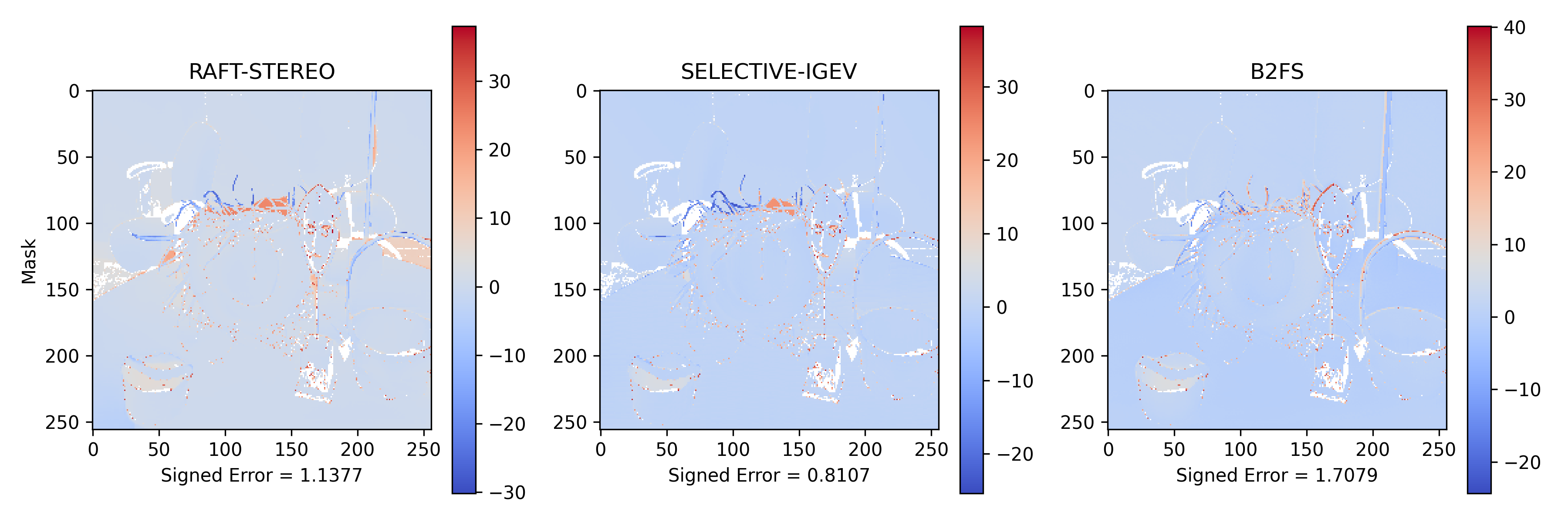}
    \caption{Mask}
    \label{fig:Mask_err}
\end{figure*}

\begin{figure*}[!ht]
    \centering
    \includegraphics[width=1.0\textwidth]{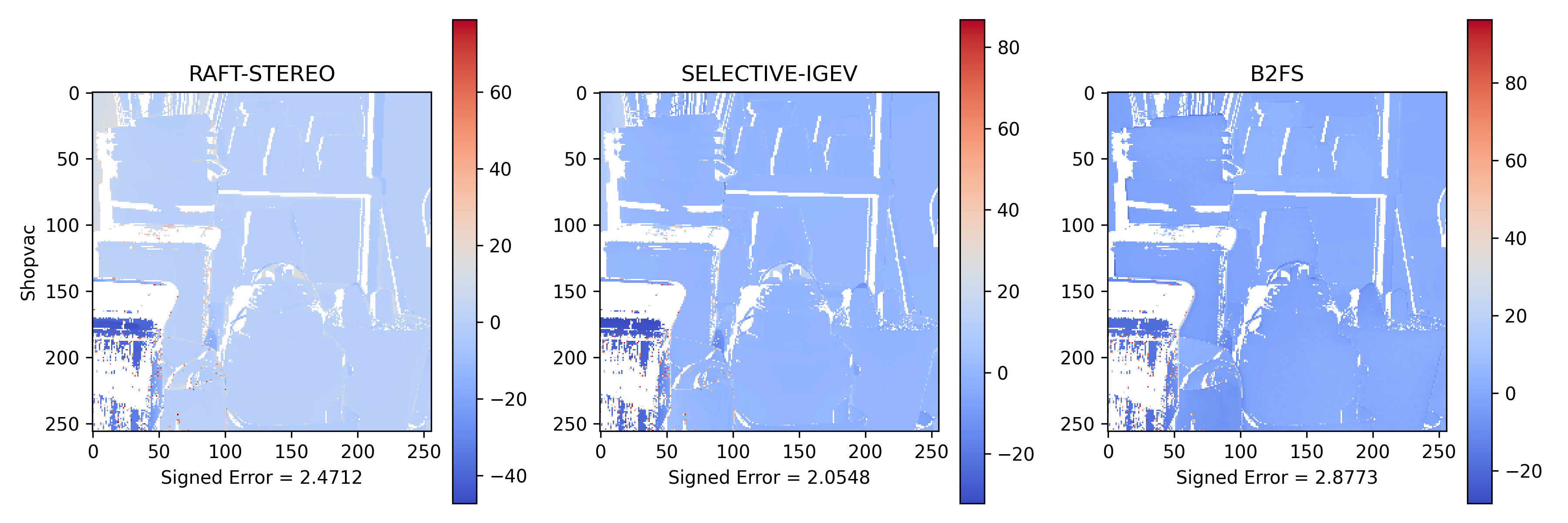}
    \caption{Shopvac}
    \label{fig:Shopvac_err}
\end{figure*}

\begin{figure*}[!ht]
    \centering
    \includegraphics[width=1.0\textwidth]{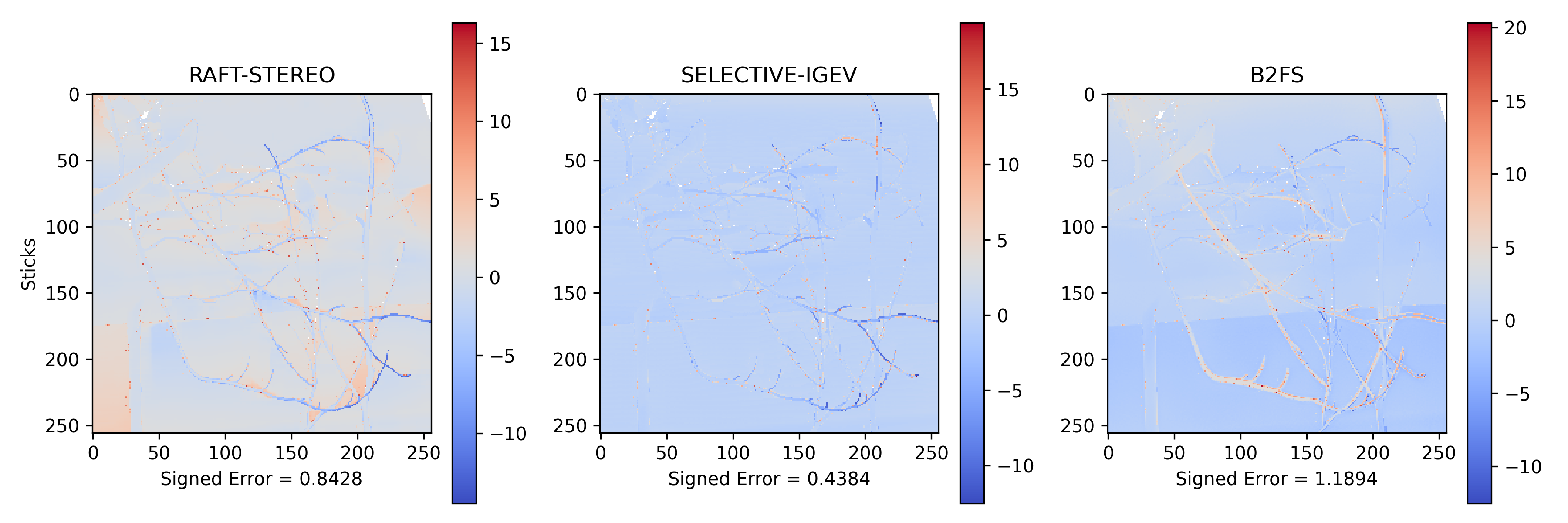}
    \caption{Sticks}
    \label{fig:Sticks_err}
\end{figure*}

\begin{figure*}[!ht]
    \centering
    \includegraphics[width=1.0\textwidth]{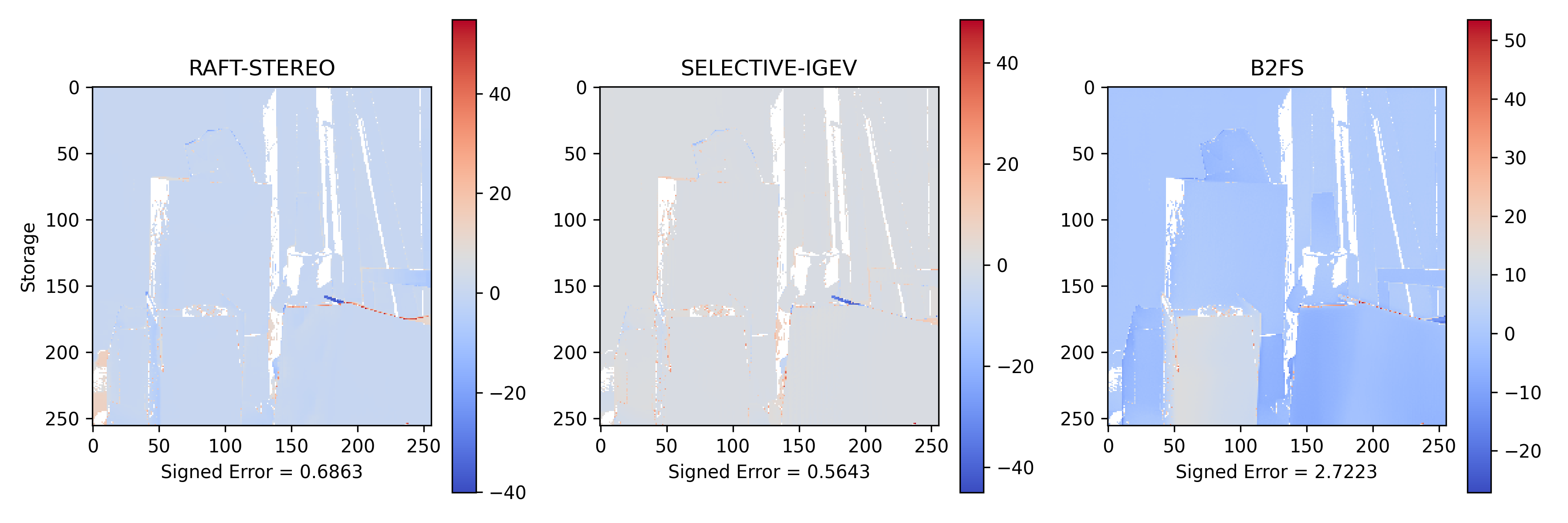}
    \caption{Storage}
    \label{fig:Storage_err}
\end{figure*}

\begin{figure*}[!ht]
    \centering
    \includegraphics[width=1.0\textwidth]{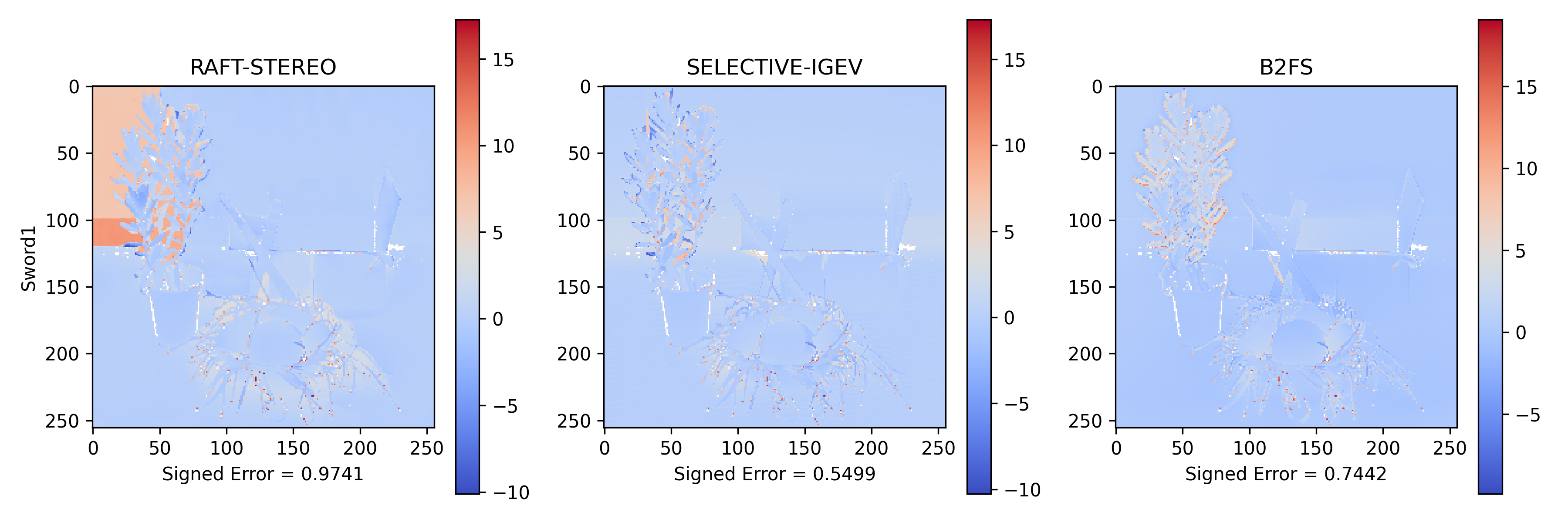}
    \caption{Sword1}
    \label{fig:Sword1_err}
\end{figure*}

\begin{figure*}[!ht]
    \centering
    \includegraphics[width=1.0\textwidth]{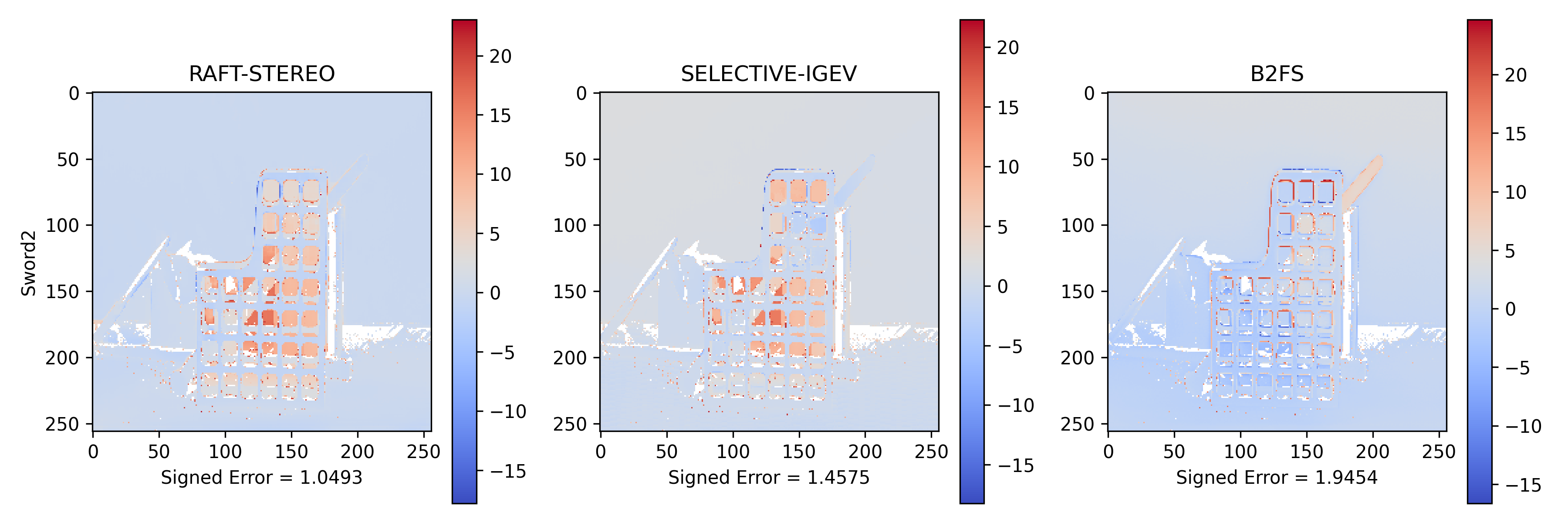}
    \caption{Sword2}
    \label{fig:Sword2_err}
\end{figure*}

\begin{figure*}[!ht]
    \centering
    \includegraphics[width=1.0\textwidth]{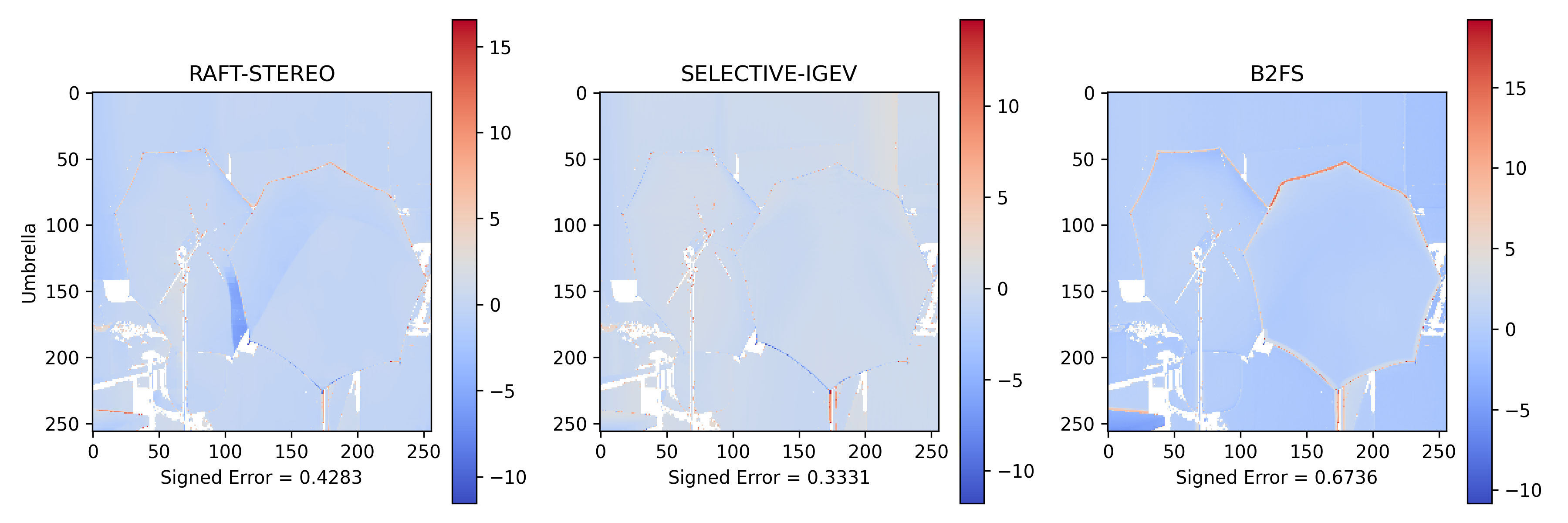}
    \caption{Umbrella}
    \label{fig:Umbrella_err}
\end{figure*}

We show the signed error for each pixel across the entire image, using red to indicate regions where the disparities should be further back (negative difference values) and blue for regions where disparities should be further ahead (positive difference values). White indicates areas where the error is not measured due to limitations in Middlebury disparity estimation. The signed error value at the bottom of each image is the average value of the absolute pixel signed error.   Although B2FS has larger overall signed error, it is typically due to a slightly larger error estimation of depth in large homogeneous regions, where feature matching was discarded. At depth discontinuities, B2FS performs significantly better. Future work includes efforts to integrate homogeneous regions feature matching to guide monocular depth solutions. See figures~\ref{fig:Backpack_err},~\ref{fig:Bicycle_err},~\ref{fig:Cable_err},~\ref{fig:Classroom1_err},~\ref{fig:Couch_err},~\ref{fig:Flowers_err},~\ref{fig:Mask_err},~\ref{fig:Shopvac_err},~\ref{fig:Sticks_err},~\ref{fig:Storage_err},~\ref{fig:Sword1_err},~\ref{fig:Sword2_err},~\ref{fig:Umbrella_err}.

\section{Impact of image resolution}
\label{sec:image_resolution}
In this section, the figures present the impact of resolution changes in the input images for RAFT-Stereo, Selective-IGEV, and B2FS. Notice that as the input image resolution decreases, B2FS preserves details, while RAFT-Stereo and Selective-IGEV struggle. We hypothesize that this effect is due to the DL algorithms being overly focused on higher-resolution images. 

\section{General Comments}
\label{sec:general-comments}
Data-driven models rely on the quality and representativeness of data to accurately capture the properties of the target domain. Currently, the most accurate stereo vision techniques are trained on thousands of synthetic and/or real image pairs to generate high-quality disparity maps. Many studies in the literature focus on creating increasingly large datasets to train deep learning (DL) models, enhancing their performance. With sufficient data coverage, zero-shot scenarios become less common as models are exposed to a vast range of situations.

\begin{figure*}[!ht]
    \centering
    \includegraphics[width=1.00\textwidth]{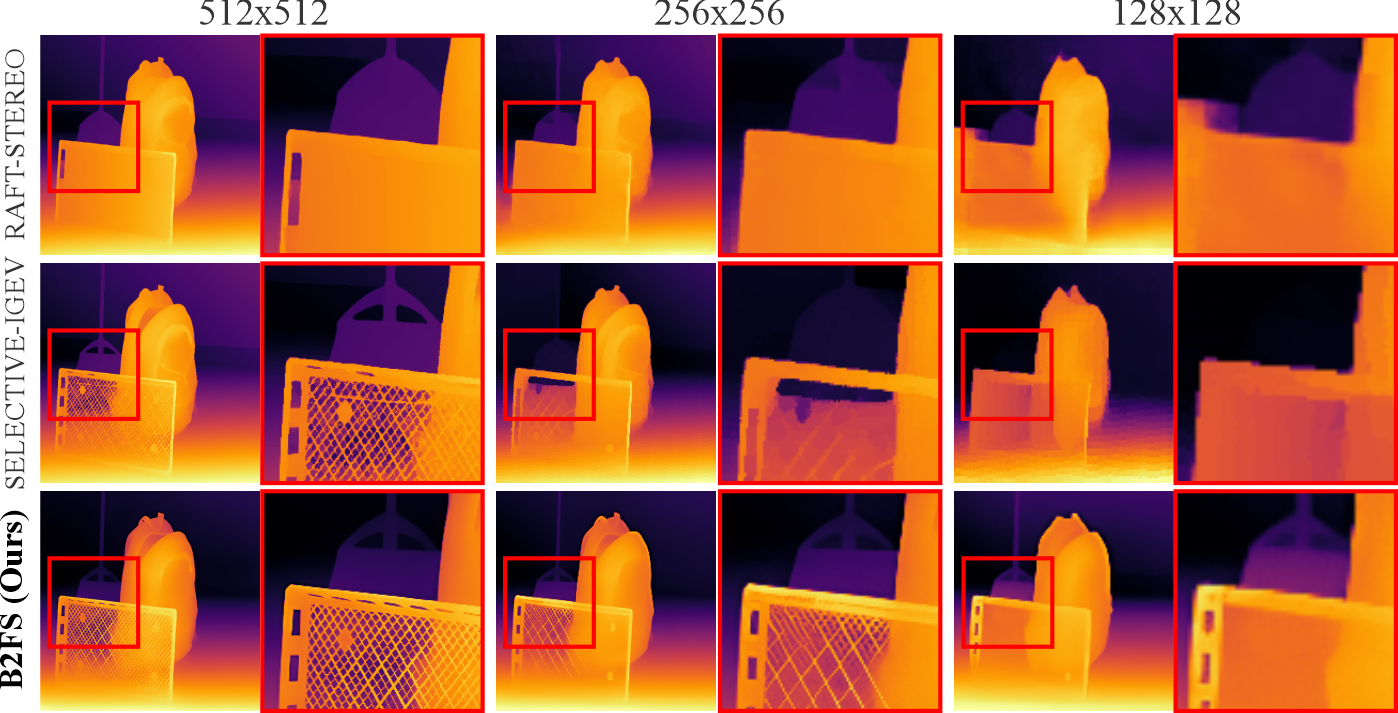}
    \caption{Backpack}
    \label{fig:Backpack_res}
\end{figure*}

\begin{figure*}[!ht]
    \centering
    \includegraphics[width=1.00\textwidth]{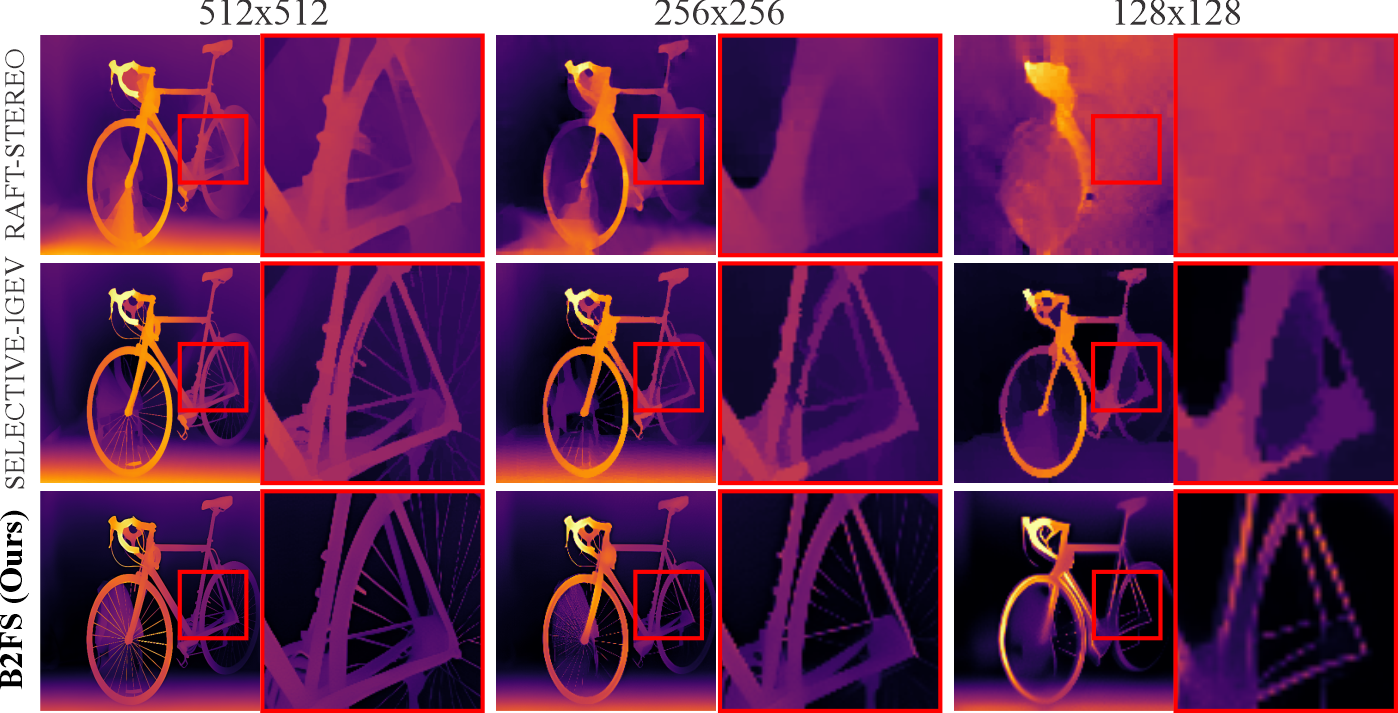}
    \caption{Bicycle}
    \label{fig:Bicycle_res}
\end{figure*}

\begin{figure*}[!ht]
    \centering
    \includegraphics[width=1.00\textwidth]{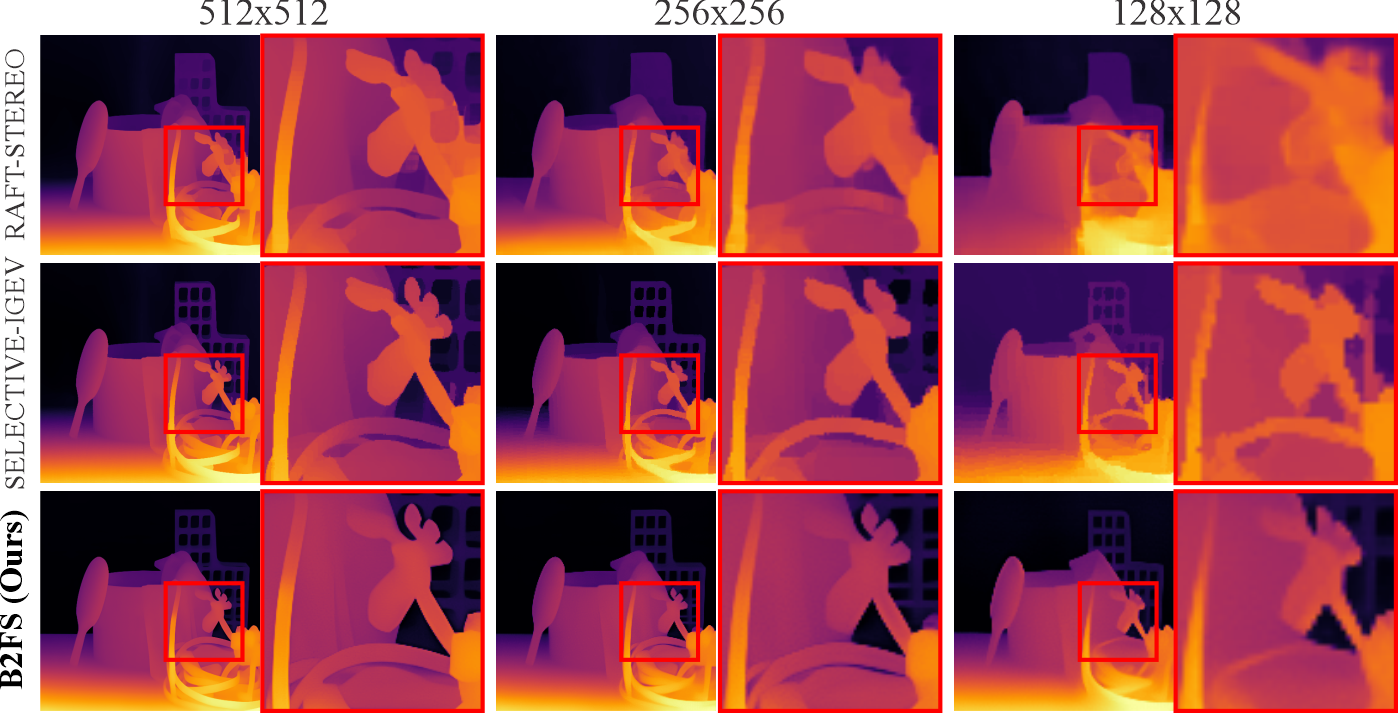}
    \caption{Cable}
    \label{fig:Cable_res}
\end{figure*}

\begin{figure*}[!ht]
    \centering
    \includegraphics[width=1.00\textwidth]{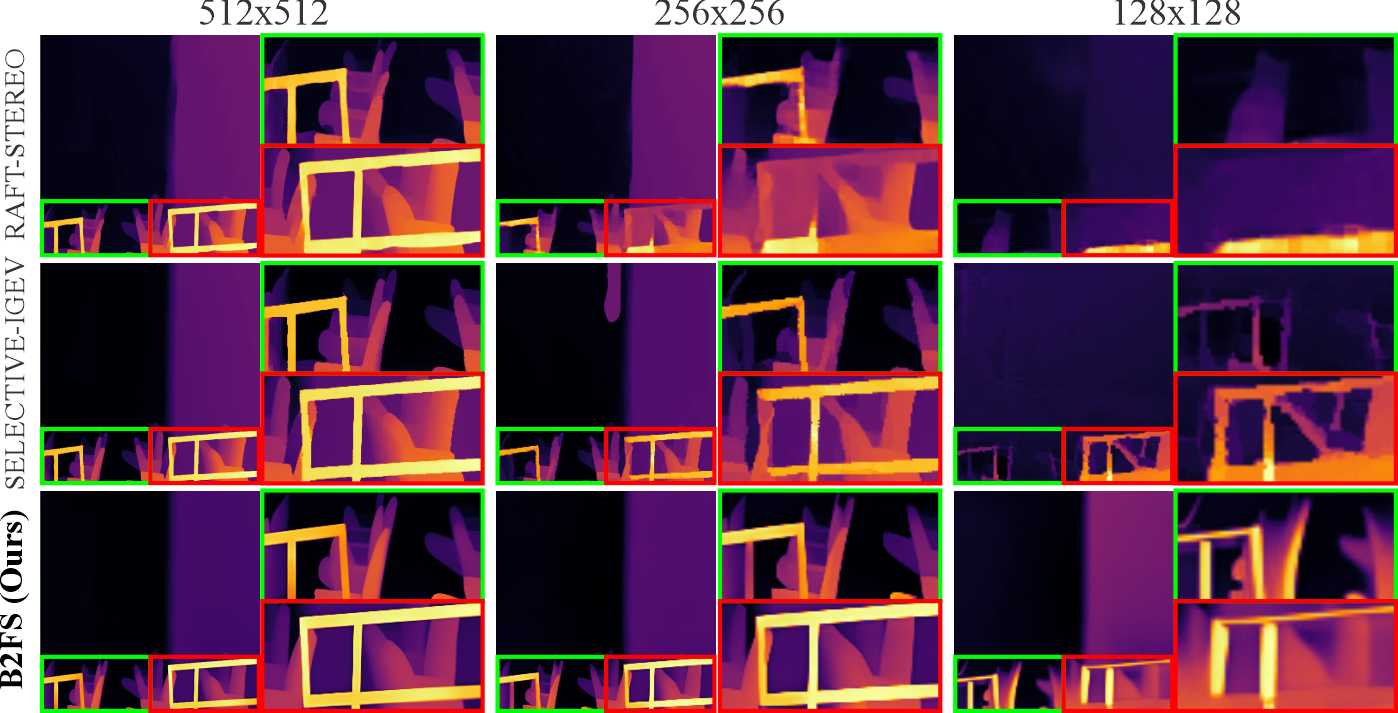}
    \caption{Classroom1}
    \label{fig:Classroom1_res}
\end{figure*}

\begin{figure*}[!ht]
    \centering
    \includegraphics[width=1.00\textwidth]{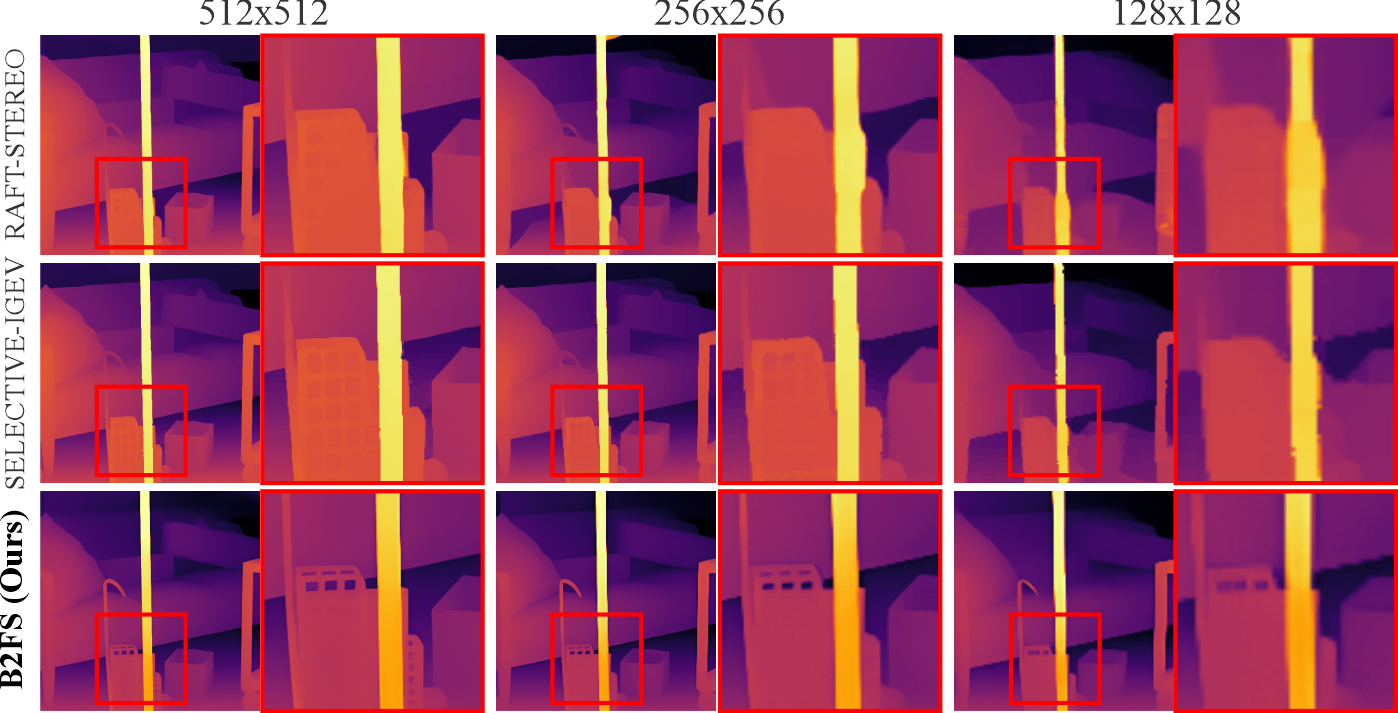}
    \caption{Couch}
    \label{fig:Couch_res}
\end{figure*}

\begin{figure*}[!ht]
    \centering
    \includegraphics[width=1.00\textwidth]{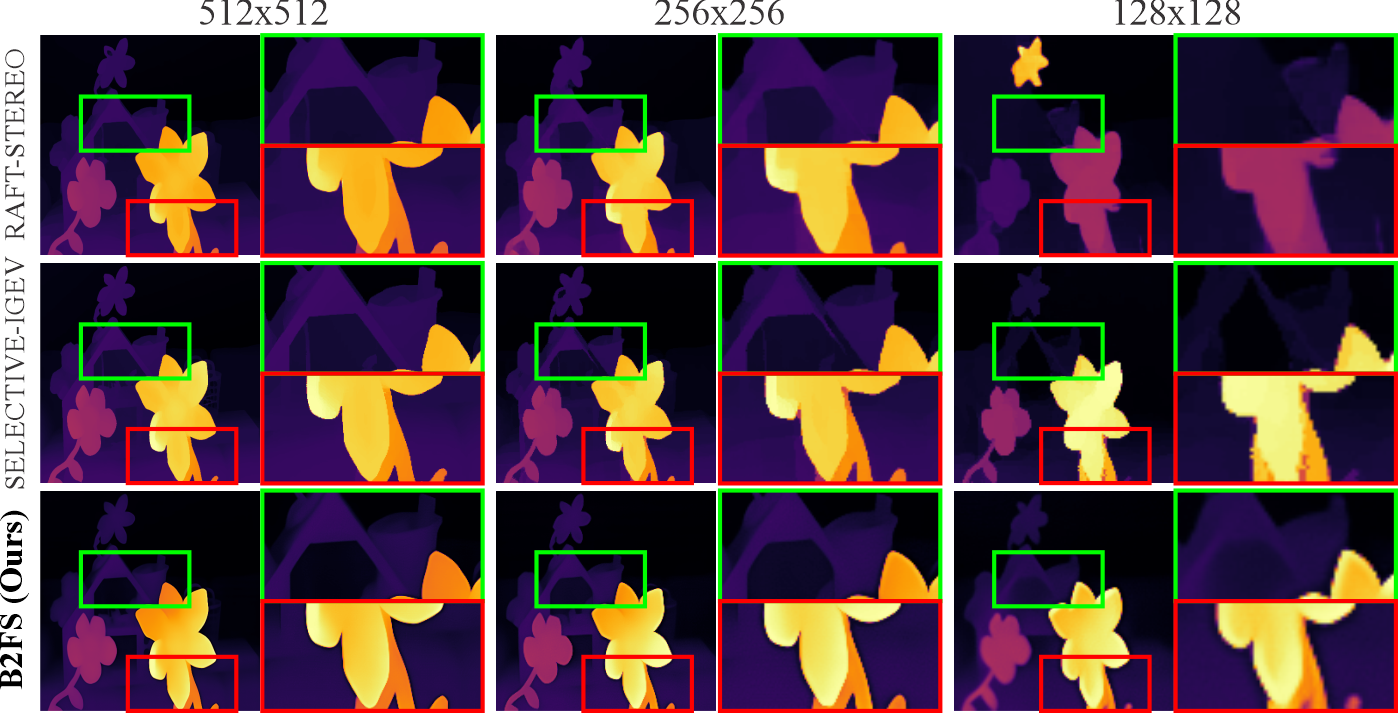}
    \caption{Flowers}
    \label{fig:Flowers_res}
\end{figure*}

\begin{figure*}[!ht]
    \centering
    \includegraphics[width=1.00\textwidth]{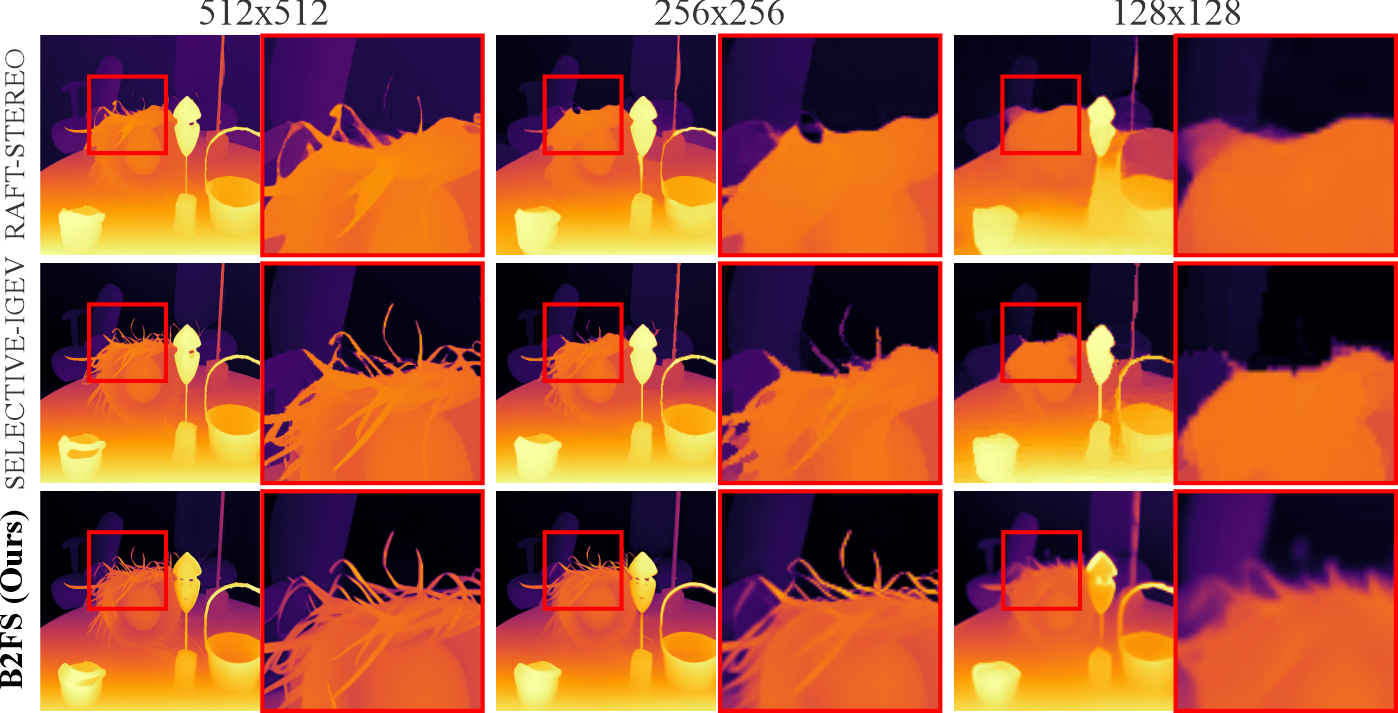}
    \caption{Mask}
    \label{fig:Mask_res}
\end{figure*}

\begin{figure*}[!ht]
    \centering
    \includegraphics[width=1.00\textwidth]{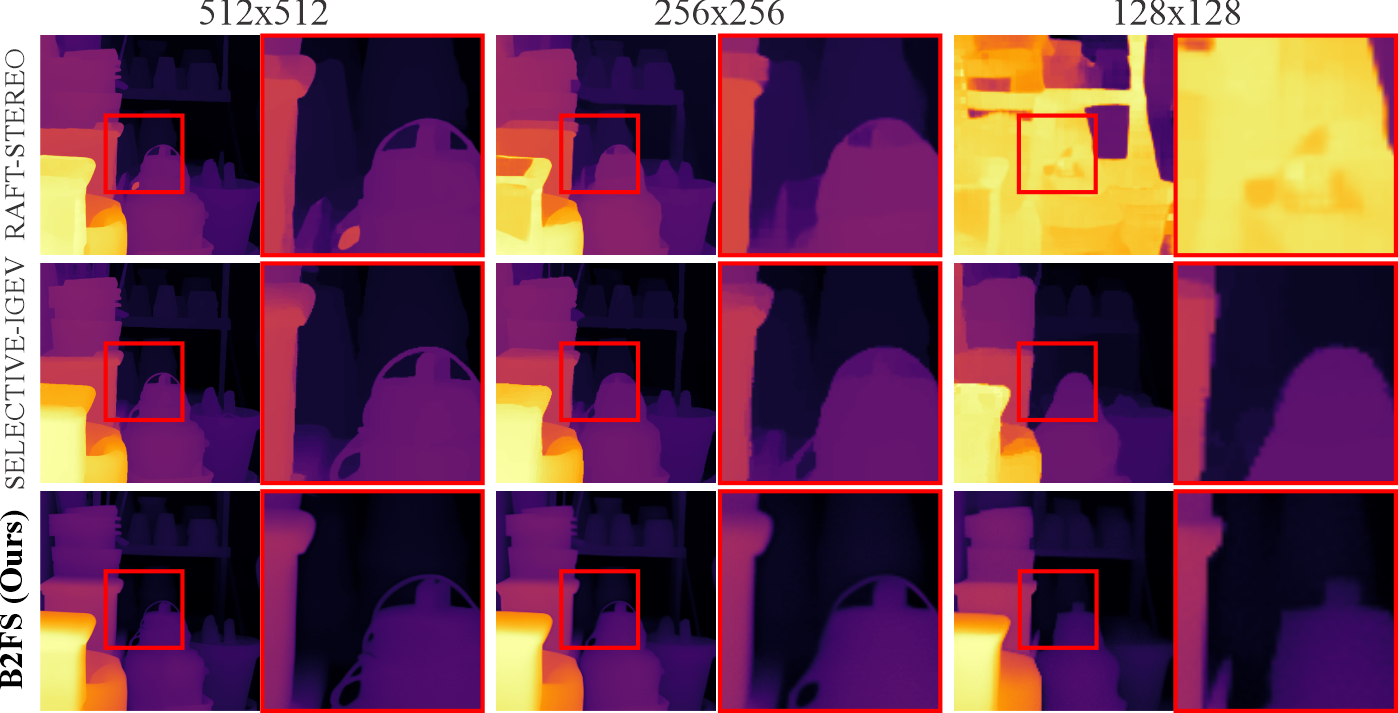}
    \caption{Shopvac}
    \label{fig:Shopvac_res}
\end{figure*}

\begin{figure*}[!ht]
    \centering
    \includegraphics[width=1.00\textwidth]{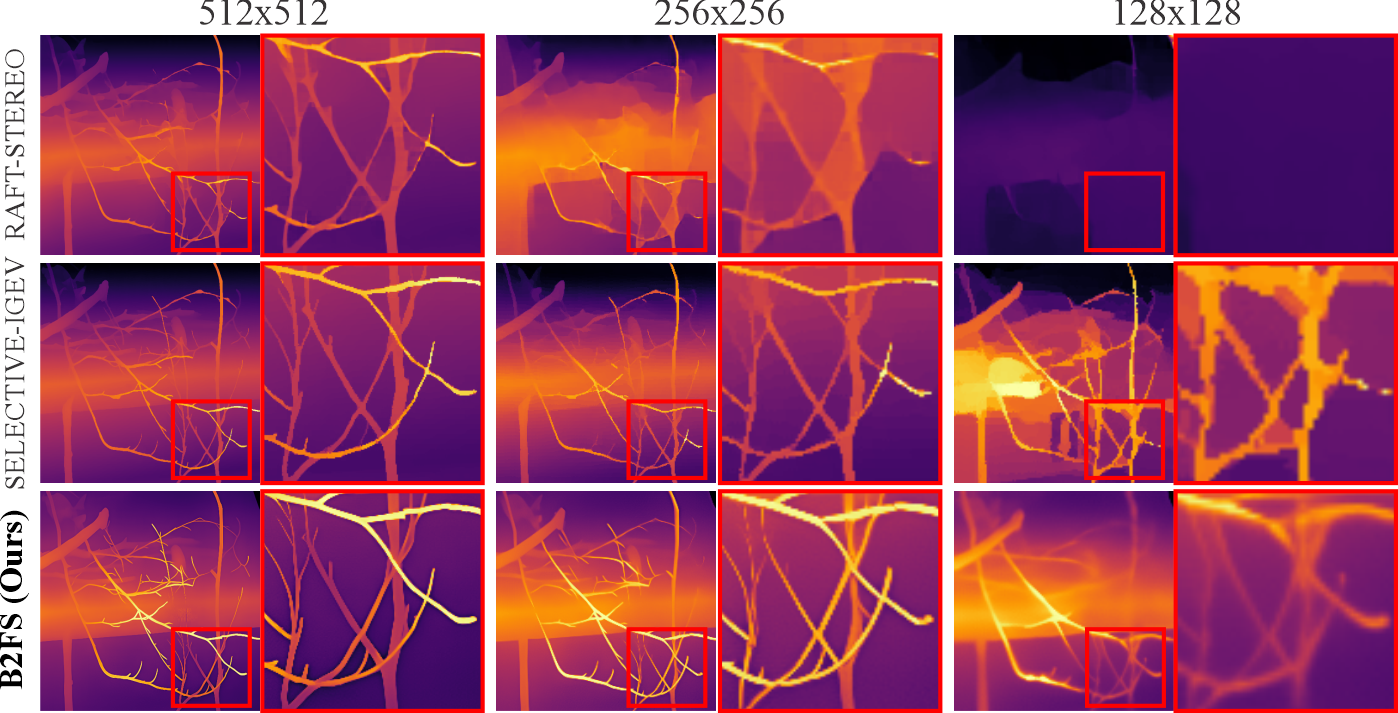}
    \caption{Sticks}
    \label{fig:Sticks_res}
\end{figure*}

\begin{figure*}[!ht]
    \centering
    \includegraphics[width=1.00\textwidth]{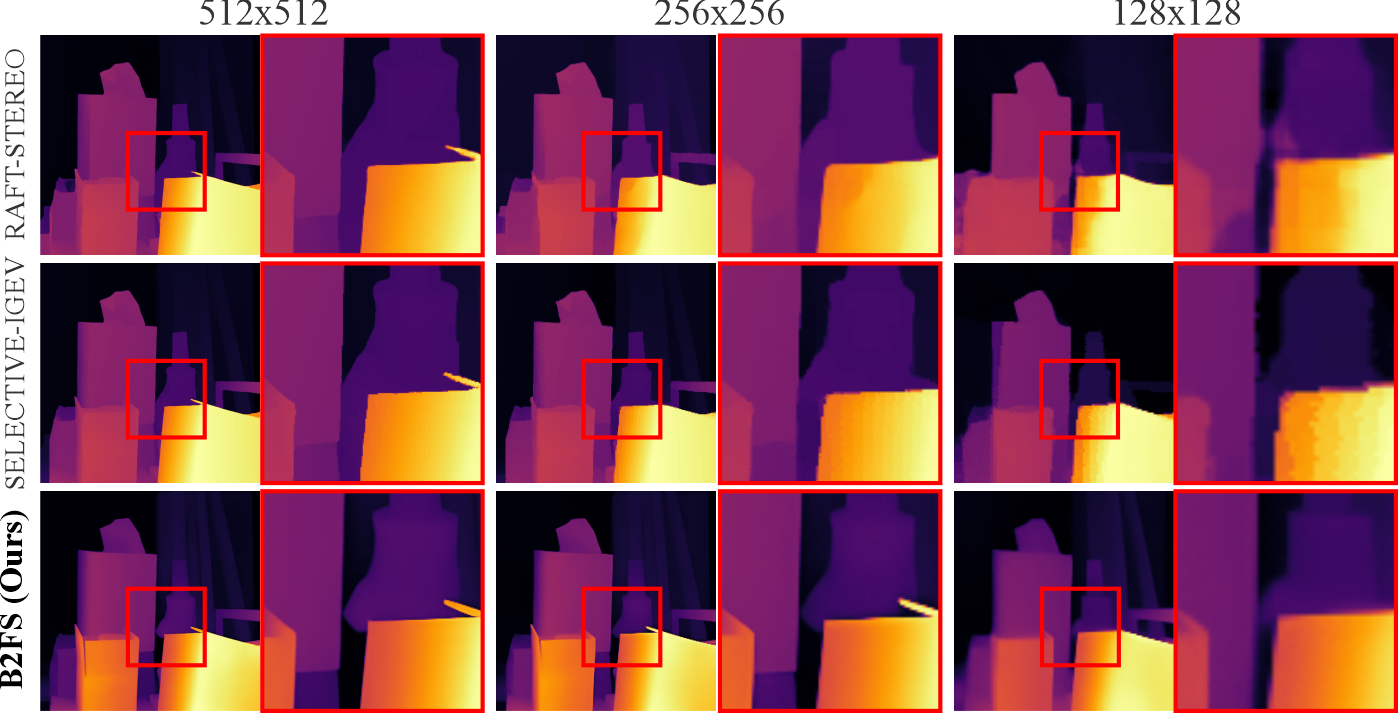}
    \caption{Storage}
    \label{fig:Storage_res}
\end{figure*}

\begin{figure*}[!ht]
    \centering
    \includegraphics[width=1.00\textwidth]{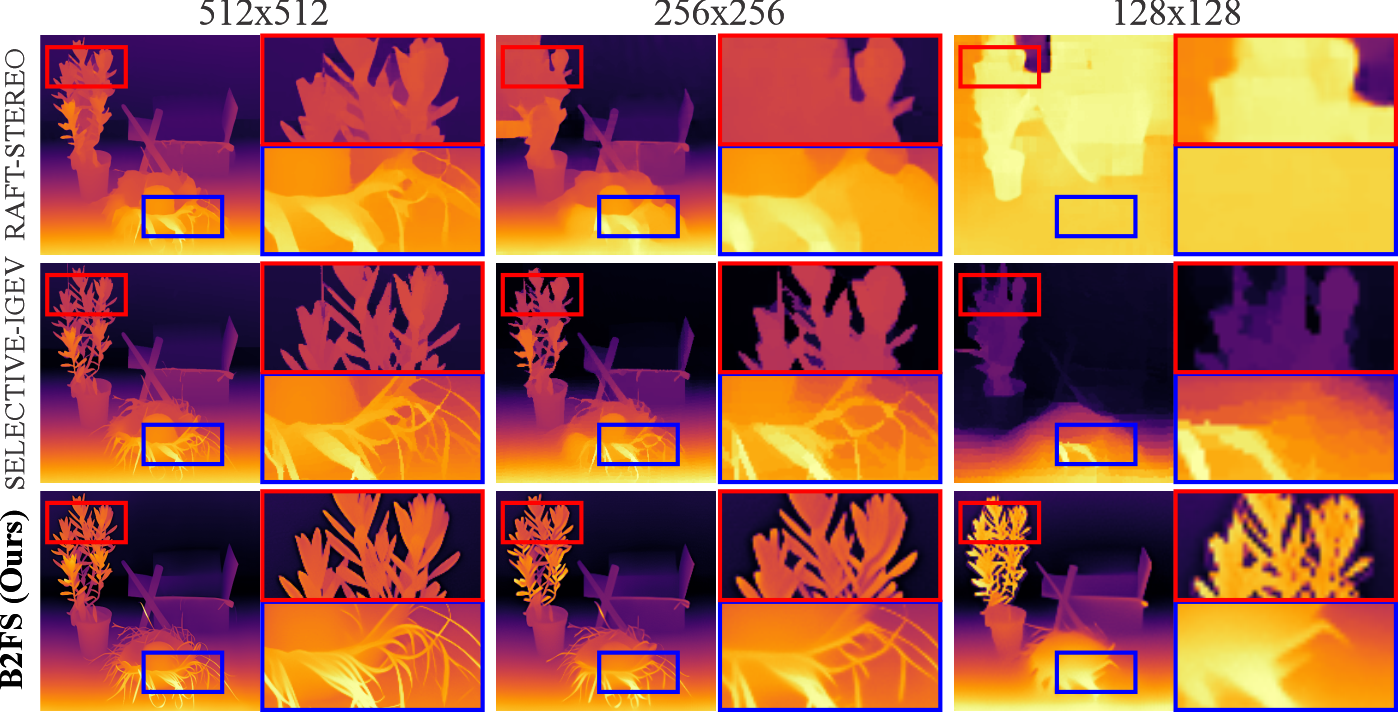}
    \caption{Sword1}
    \label{fig:Sword1_res}
\end{figure*}

\begin{figure*}[!ht]
    \centering
    \includegraphics[width=1.00\textwidth]{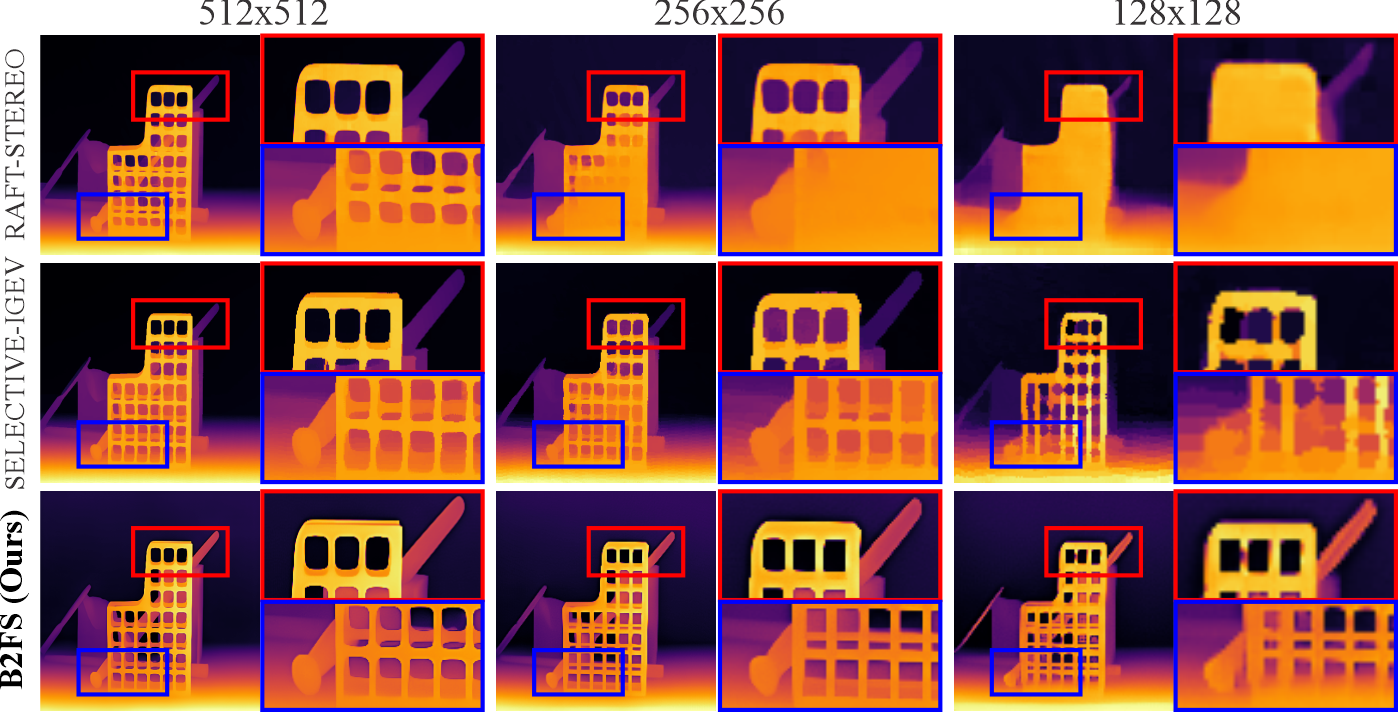}
    \caption{Sword2}
    \label{fig:Sword2_res}
\end{figure*}

\begin{figure*}[!ht]
    \centering
    \includegraphics[width=1.00\textwidth]{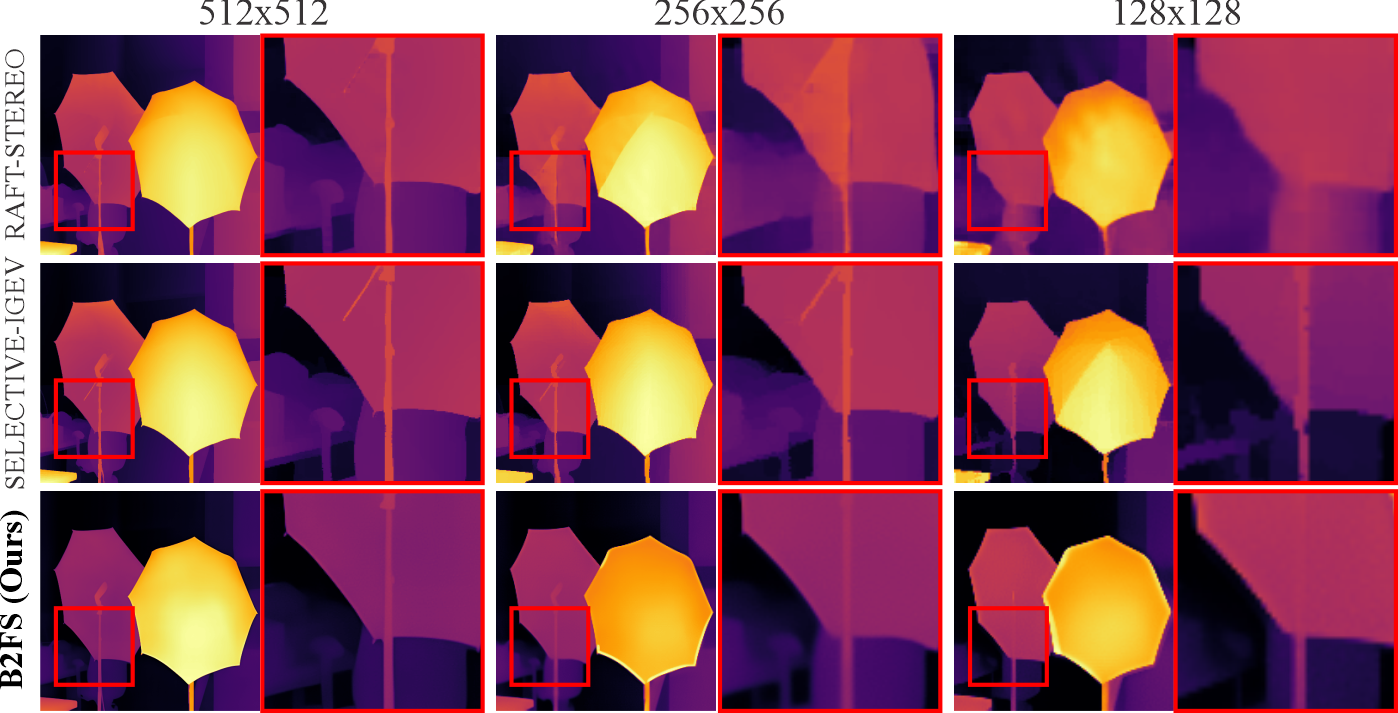}
    \caption{Umbrella}
    \label{fig:Umbrella_res}
\end{figure*}

While expanding datasets is essential for faithfully capturing real-world patterns, it is not the only step necessary for understanding them. It is important to emphasize that purely data-driven models can also incorporate a reasoning step based on world priors related to the task. This can lead to more general results and simpler models, going beyond data fitting. See figures~\ref{fig:Backpack_res},~\ref{fig:Bicycle_res},~\ref{fig:Cable_res},~\ref{fig:Classroom1_res},~\ref{fig:Couch_res},~\ref{fig:Flowers_res},~\ref{fig:Mask_res},~\ref{fig:Shopvac_res},~\ref{fig:Sticks_res},~\ref{fig:Storage_res},~\ref{fig:Sword1_res},~\ref{fig:Sword2_res},~\ref{fig:Umbrella_res}.

\end{document}